\documentclass[12pt]{article}

\usepackage{scicite}
\usepackage{times}
\usepackage[utf8]{inputenc} 
\usepackage[T1]{fontenc}    
\usepackage{url}            
\usepackage{booktabs}       
\usepackage{nicefrac}       
\usepackage{microtype}      
\usepackage{graphicx,color}
\graphicspath{ {./ArXiv_Figure/} }
\usepackage[labelfont=bf]{caption}
\DeclareGraphicsExtensions{.pdf}
\usepackage{booktabs}

\usepackage{algorithm}
\usepackage{algorithmic}
\usepackage{enumitem}
\usepackage{typed-checklist}

\usepackage{multirow}
\usepackage{multicol}
\usepackage{amsfonts}       
\usepackage{amsmath}       
\usepackage{amssymb}

\usepackage{pdfpages}

\newcommand{\Fig}[1]{Figure~\ref{#1}}  
\newcommand{\FigS}[2]{Figure.~\ref{#1}\,#2}
\newcommand{\fig}[1]{Fig.~\ref{#1}}    
\newcommand{\figS}[2]{Fig.~\ref{#1}\,#2}    
\newcommand{\tab}[1]{Table~\ref{#1}}
\newcommand{\vid}[1]{Video~#1}

\newcommand{\eqn}[1]{Eq.~\ref{#1}} 
\renewcommand{\sec}[1]{Sec.~\ref{#1}} 
\newcommand{\supp}[1]{Suppl.~\ref{#1}}
\usepackage{xspace}
\newcommand{\method}{\textit{Insight}\xspace}

\usepackage{xr}
\makeatletter
\newcommand*{\addFileDependency}[1]{
  \typeout{(#1)}
  \@addtofilelist{#1}
  \IfFileExists{#1}{}{\typeout{No file #1.}}
}
\makeatother

\makeatletter
\DeclareRobustCommand\onedot{\futurelet\@let@token\@onedot}
\def\@onedot{\ifx\@let@token.\else.\null\fi\xspace}
\def\eg{e.g\onedot}
\def\ie{i.e\onedot}

\makeatother

\usepackage{xcolor}
\definecolor{ourblue}{rgb}{0.368,0.507,0.71}
\definecolor{ourorange}{rgb}{0.881,0.611,0.142}
\definecolor{ourgreen}{rgb}{0.56,0.692,0.195}
\definecolor{ourred}{rgb}{0.923,0.386,0.209}
\definecolor{ourviolet}{rgb}{0.528,0.471,0.701}
\definecolor{ourbrown}{rgb}{0.772,0.432,0.102}
\definecolor{ourlightblue}{rgb}{0.364,0.619,0.782}
\definecolor{ourdarkgreen}{rgb}{0.572,0.586,0.}

\usepackage[colorlinks=true, linkcolor=ourblue, citecolor=ourgreen,urlcolor=ourblue]{hyperref}

\title{A soft thumb-sized vision-based sensor with\\ accurate all-round force perception}

\author{Huanbo Sun$^{1\ast}$, Katherine J. Kuchenbecker$^{2}$, and Georg Martius$^{1\ast}$\\
        \normalsize{$^{1}$ Autonomous Learning Group, Max Planck Institute for Intelligent Systems,}\\[-.2em]
        \normalsize{T\"ubingen, Germany.}\\
        \normalsize{$^{2}$ Haptic Intelligence Department, Max Planck Institute for Intelligent Systems,}\\[-.2em]
        \normalsize{Stuttgart, Germany.}\\
        \normalsize{$^\ast$ Corresponding author. Emails: \{huanbo.sun$\mid$georg.martius\}@tuebingen.mpg.de}}

\begin{document}
\maketitle
\begin{abstract} 
Vision-based haptic sensors have emerged as a promising approach to robotic touch due to affordable high-resolution cameras and successful computer-vision techniques.
However, their physical design and the information they provide do not yet meet the requirements of real applications.
We present a robust, soft, low-cost, vision-based, thumb-sized 3D haptic sensor named \method:
it continually provides a directional force-distribution map over its entire conical sensing surface.
Constructed around an internal monocular camera, the sensor has only a single layer of elastomer over-molded on a stiff frame to guarantee sensitivity, robustness, and soft contact.
Furthermore, \method{} is the first system to combine photometric stereo and structured light using a collimator to detect the 3D deformation of its easily replaceable flexible outer shell.
The force information is inferred by a deep neural network that maps images to the spatial distribution of 3D contact force (normal and shear).
\method{} has an overall spatial resolution of 0.4\,mm, force magnitude accuracy around 0.03\,N, and force direction accuracy around 5 degrees over a range of 0.03--2\,N for numerous distinct contacts with varying contact area.
The presented hardware and software design concepts can be transferred to a wide variety of robot parts.
\end{abstract}

\section{Introduction}
Robots have the potential to perform useful physical tasks in a wide range of application areas~\cite{Multidrone,Quadrupedal,GraspPlanning,HRI}.
To robustly manipulate objects in complex and changing environments, a robot must be able to perceive when, where, and how its body is contacting other things.
Although widely studied and highly successful for environment perception at a distance, centrally mounted cameras and computer vision are poorly suited to real-world robot contact perception due to occlusion and the small scale of the deformations involved.
Instead, robots need touch-sensitive skin, but few haptic sensors exist that are suitable for practical applications.

Recent developments have shown that machine-learning-based approaches are especially promising for creating dexterous robots~\cite{OpenAI,Nagabandi2019:DexManipul,Quadrupedal}.
In such self-learning scenarios and real-world applications, the need for extensive data makes it particularly critical that sensors are robust and keep providing good readings over thousands of hours of rough interaction.
Importantly, machine learning also opens new possibilities for tackling this haptic sensing challenge by replacing handcrafted numeric calibration procedures with end-to-end mappings learned from data~\cite{Machinelearning}.

There have been many efforts to create haptic sensors~\cite{Review} that can quantify contact across the surface of a robot: previous successful designs produced measurements using resistive~\cite{BioTac,Resistive,Piezoresistive,HapDefX,ResistiveTaunyazov-RSS-20}, capacitive~\cite{e-Skin,HexoSkin,HexoSkin-Humanoids}, ferroelectric~\cite{Ferroelectric}, triboelectric~\cite{Triboelectric}, and optoresistive~\cite{Optoresistive,BaiSLIMS} transduction approaches.
More recently, vision-based haptic sensors~\cite{GelSight,SlimFEM,Beads,L-Skeleton,TacTip,Eletroluminescent} have demonstrated a new family of solutions, typically using an internal camera that views the soft contact surface from within.
However, these existing sensors tend to be fragile, bulky, insensitive, inaccurate, and/or expensive.
By considering the goals and constraints from a fresh perspective, we have invented a vision-based sensor that overcomes these challenges and is thus suitable for robotic dexterous manipulation.

A detailed comparison of representative state-of-the-art sensors is shown in \tab{tab:Overview}. We highlight the most important differences in the following and refer the reader to the Methods section for a more thorough examination.
The mechanical designs of all previously developed sensors employ multiple functional layers, which are complex to fabricate and can be delicate. \method is the only sensor with a single soft layer.
Many tasks benefit from a large 3D sensing surface rather than small 2D sensing patches; however, only a few other sensors also offer 3D surfaces \cite{OmniTact,GelTip,TacTip,Romero2020GelSight2.5}.
Their design is often technically complex, for instance, with multiple cameras \cite{OmniTact}
or special lenses \cite{TacTip}, whereas \method needs only a single camera and simple manufacturing techniques.
Depending on their mechanical design, sensors also have widely varying sensing surface area and sensor volume.
We provide \emph{area per volume} (A/V) in \tab{tab:Overview} as a measure of compactness and find that \method is the most compact vision-based sensor with the largest sensing surface.

Existing sensors also differ in the type of information they provide. Most sensors provide only localization of a single contact \cite{LeeCapacitive, BaiSLIMS,OmniTact,GelTip,TacTip}, and some sensors additionally provide a force magnitude \cite{BioTac,YanHallEffect,Beads} without force direction.
Others are specialized for measuring contact area shape \cite{GelSight,DIGIT,Romero2020GelSight2.5}.
Although real contacts will be multiple and complex, a spatially extended map of 3D contact forces over the surface, which we call a \emph{force map}, is only rarely provided, e.g., \cite{SlimFEM}.
\method is the only sensor that provides a force map across a 3D surface so a robot can have detailed directional information about simultaneous contacts.
Many sensors rely on analytical data processing \cite{GelTip,TacTip,GelForce,SlimFEM}, which requires careful calibration; it is difficult to obtain correct force amplitudes with such an approach because materials are often inhomogeneous and the assumption of linearity between deformation and force is often violated.
Data-driven approaches like those used with a BioTac \cite{BioTac}, GelSight\cite{GelSight}, OmniTact \cite{OmniTact}, and \method can deal with these problems but naturally require copious good data.

In this paper, we present a new soft thumb-sized sensor with all-around force-sensing capabilities enabled by vision and machine learning; it is durable, compact, sensitive, accurate, and affordable (less than $100\,$USD).
Because it consists of a flexible shell around a vision sensor, we name it \method{}.
We initially designed the sensor for dexterous manipulation devices with behavioral learning scenarios in mind.
However, our sensor is suitable for many other applications, and our technology can be adapted to create a variety of differently shaped 3D haptic sensing systems.

\begin{figure}
    \centering
    \vspace*{-2cm}
    \includegraphics[width=0.95\textwidth]{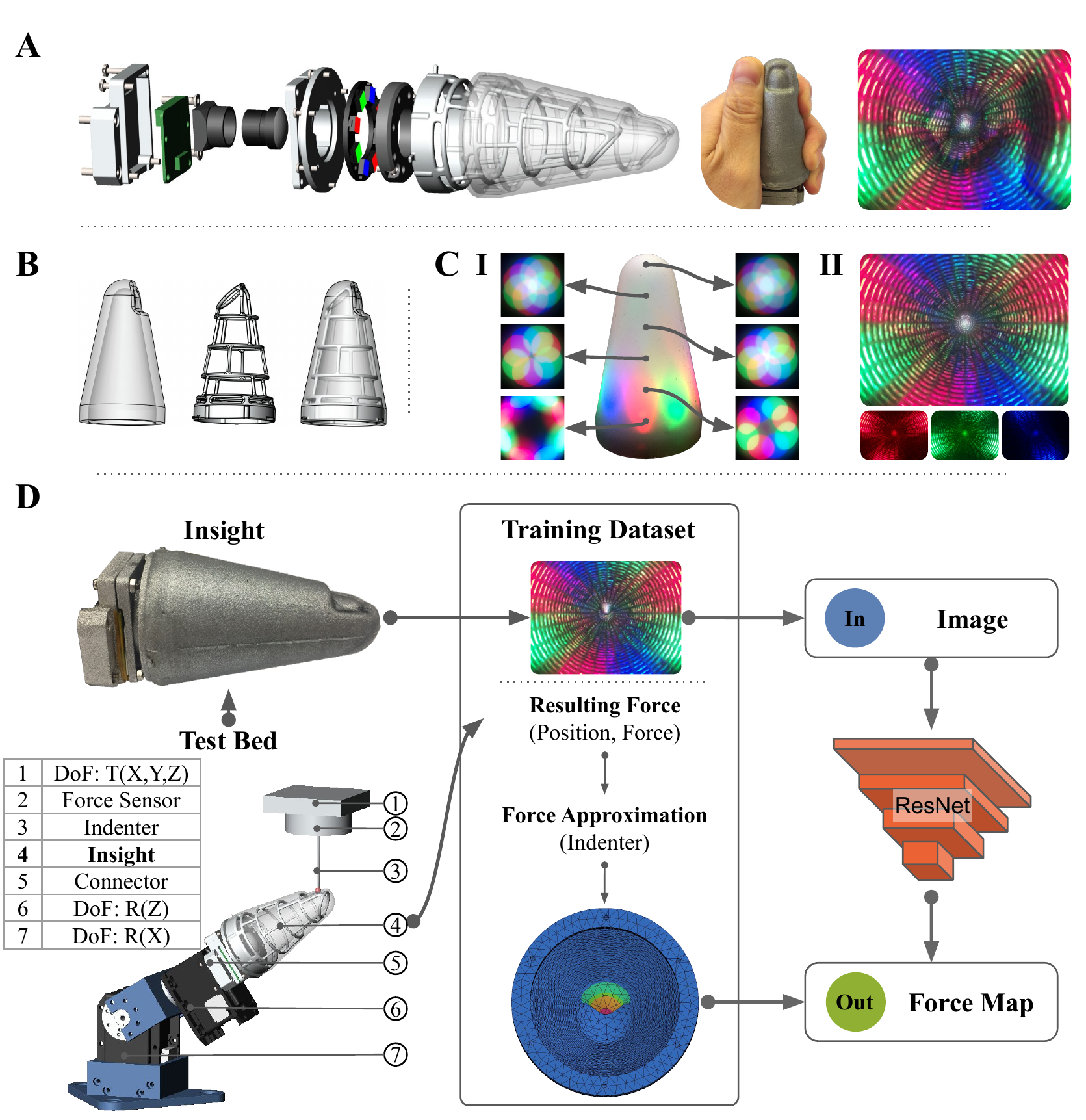}
    \caption{\textbf{The design principles of \method{}}.
    \textbf{A} depicts the overall structure of the sensor with its hybrid mechanical construction and internal imaging system.
    For comparison, the sensor is shown in a human hand, next to the corresponding camera view.
    \textbf{B} shows the pure elastomer (left), the stiff hollow skeleton (middle), and both over-molded together (right).
    \textbf{C-I} illustrates the internal lighting using a translucent shell: the LED ring with apertures creates light cones, visualized by their projections on flat horizontal planes.
    \textbf{C-II} depicts light projection patterns within as seen by the camera in the undeformed opaque shell.
    \textbf{D} presents the data processing pipeline.
    The machine-learning model is trained on data collected by an automatic test bed.
    Each data point combines one image from the camera with the indenter's contact position and orientation, contact force vector, and diameter, which are used to calculate a ground-truth force distribution map from an approximate model under consistent contact forces.}
    \label{fig:Figure1}
\end{figure}

\Fig{fig:Figure1} shows the principles behind the design of \method{}.
The skin is made of a \textbf{soft elastomer} over-molded~\cite{Overmolding} on a hollow \textbf{stiff skeleton} to maintain the sensor's shape and allow for high interaction forces without damage (\figS{fig:Figure1}{B}).
It utilizes \textbf{shading} effects~\cite{PS} and \textbf{structured light}~\cite{SL} to monitor the 3D deformation of the sensing surface with a single \textbf{camera} from the inside (\figS{fig:Figure1}{C}).
The sensor's output is computed by a data-driven \textbf{machine-learning} approach~\cite{HapDefX,Resistive}, which directly infers distributed contact-force information from raw camera readings, avoiding complicated calibration or any hand-crafted post-processing steps (\figS{fig:Figure1}{D}).

\method{} is evaluated against several rigorous performance criteria within this paper.
When indented by a hemispherical tip with a diameter of $4\,$mm at a force amplitude up to $2.0\,$N, the sensor can achieve an average localization accuracy around $0.4\,$mm and a force accuracy around $0.03\,$N.
By directly estimating both the normal and shear components of each applied force vector, the sensor reaches an average directional estimation error around $5^\circ$.
Moreover, in the absence of contact, \method{} is sensitive enough to recognize its posture relative to gravity based only on the deformations caused by its own self-weight, which are not detectable by eye.

\begin{table*}
    \centering
    \caption{A comparison between state-of-the-art haptic sensors and our design. An upward arrow ($\uparrow$) indicates that higher values are better, while a downward arrow ($\downarrow$) means lower is better.}
    \resizebox{\textwidth}{!}{\begin{tabular}{@{\ }ll@{}c@{\ \ }c ccc@{\ }c l@{\ \ }c@{\ \ }c@{\ \ }c@{}c@{\ }}
        \toprule
        \multirow{2}{*}{Sensor Name} &Transduction &\# of &Surface &Area $\uparrow$ &A/V $\uparrow$ &Data &Output & \multicolumn{4}{|c|}{Sensing Error $\downarrow$} & \multirow{2}{*}{Notes} \\
        &Method &Layers $\downarrow$ &Shape $\uparrow$ &[mm$^2$] &[mm$^{-1}$] &Processing &Format& \multicolumn{1}{|c}{$P$\,[mm]} & $F_n$\,[N]& $F_{s}$\,[N] & \multicolumn{1}{c|}{$\alpha$\,[$^\circ$]} & \\
        \midrule
        BioTac                     & Resistive          & 2          & Half 3D     & 484           & 0.060          & FCN             & Location+Force     & 1.4~\cite{BioTac1} & 0.85                    & $\sim$0.48~\cite{BioTac} & 10~\cite{BioTac2} & costly and delicate \\
        Lee et al.~\cite{LeeCapacitive} & Capacitive         & 5          & 2D          & 484           & 1.000          & ---             & Location           & 2                  & ---                     & ---                     & ---                     & 10 mN full-scale range               \\
        Yan et al.~\cite{YanHallEffect} & Hall Effect        & 2          & 2D          & 324           & 0.179          & FCN             & Location+Force     & 0.1                & 0.15                    & ---                     & ---                     & blind to shear force                 \\
        SLIMS~\cite{BaiSLIMS}      & Optical            & 3          & 1D          & 350           & 0.050          & ---             & Location           & 10                 & ---                     & ---                     & ---                     & sensitive to ambient light           \\
        \midrule
        GelSight~\cite{GelSight}   & Cam+PS+Markers      & 4          & 2D          & 250           & 0.003          & CNN             & Shape+Force        & ---                & 0.67                    & $\sim$0.17               & ---                     & complex to manufacture               \\
        GelSlim~\cite{SlimFEM}    & Cam+PS+Markers      & 5         & 2D          & 1200          & 0.006          & inv.FEM            & Force Map          & ---                & 0.32                    & $\sim$0.22               & ---                     & tears after 1500 contacts             \\
        OmniTact~\cite{OmniTact}   & 5$\times$Cam+PS & 2          & 3D          & 3110          & 0.083          & ResNet          & Location      & 0.4                & ---                     & ---                     & ---                     & 5 cameras, ambient light             \\
        GelTip~\cite{GelTip}       & Cam+PS             & 2          & 3D          & 2513          & 0.052          & Numeric         & Location           & 5                  & ---                     & ---                     & ---                     & imaging artifacts on tip             \\
        TacTip~\cite{TacTip}       & Cam+Markers         & 3          & 3D          & 2500          & 0.025          & Numeric         & Location           & 0.2                & ---                     & ---                     & ---                     & marker density limits res.         \\
        Sf. \& D'A.~\cite{Beads}   & Cam+Beads          & 3          & 2D          & 900           & 0.008          & FCN             & Location+Force     & 0.2                & 0.05                    & ---                     & ---                     & heavy, blind to shear             \\
        Sf. \& D'A.~\cite{Beads2}   & Cam+Beads          & 3          & 2D          & 900           & 0.008          & CNN             & Force Map     & ---                & 0.13                    & 0.05                     & ---                     & no real shear experiment             \\
        DIGIT~\cite{DIGIT}         & Cam+PS+Markers      & 3          & 2D          & 304           & 0.031          & ResNet          & Shape              & ---                & ---                     & ---                     & ---                     & difficult to extend to 3D           \\
        Romero et al.~\cite{Romero2020GelSight2.5}    & Cam+PS & 2 & Half 3D & 2069 & 0.084 & Numeric &Shape & --- & --- & --- & --- & limited sensing area\\
        \textbf{\method{}}           & \textbf{Cam+PS+SL} & \textbf{1} & \textbf{3D} & \textbf{4800} & \textbf{0.088} & \textbf{ResNet} & \textbf{Force Map} & \textbf{0.4}       & \textbf{0.03}           &\textbf{0.03}              & \textbf{5}                 & \textbf{this paper}                  \\
        \bottomrule
    \end{tabular}}\label{tab:Overview}
\end{table*}

\section{Results}
\subsection*{Principles of operation and design}
At the core of our design is a single camera that observes the sensor's opaque over-molded elastic shell from the inside (\figS{fig:Figure1}{A}).
Photometric effects and structured lighting enable it to detect the tiny deformations of the sensor surface that are caused by physical contact.
In principle, the contact force vectors could be numerically computed from the observed deformations according to elastic theory, but the material properties are not uniform, and the necessary assumption of a linear relationship between deformation and force~\cite{GelForce,TheoryofElastics} is often violated.
In our approach, machine learning greatly simplifies this process by mapping images directly to force distribution maps.
The details are shown in \fig{fig:Figure1} and explained in the following.

\paragraph{Mechanics}
We aim at a compliant and sensitive sensing surface because of the favorable properties of soft materials for manipulating objects~\cite{SoftRobo-IJRR}, for safer interactions in human environments~\cite{SoftRobo-Front}, and to limit the instantaneous impact forces caused by unforeseen collisions in robotic systems~\cite{SoftRobo-IEEE}.
Nevertheless, the direct application of soft materials in sensor design is nontrivial because they cannot withstand larger interaction forces.
If thin structures are formed from a material with low Young's modulus, even gravity and inertial effects change their shape considerably~\cite{SoftRobo-Nature}.

To ensure a compliant sensing surface, high contact sensitivity, and robustness against self-motion, we design a soft-stiff hybrid structure using over-molding (\figS{fig:Figure1}{B})~\cite{Overmolding}.
The structure is composed of two parts.
One is a flexible elastomer (Young's modulus around $70\,$kPa, hardness around $00-30$ in Shore-00) to sense the contact, and the other is a skeleton made of aluminum alloy (Young's modulus around $70\,$GPa) to support the sensing surface.
In this way, the sensor is not only structurally stable so it keeps its overall shape under high contact forces but also sensitive so that gentle interaction forces cause local deformations.
In contrast to our approach, other successful curved vision-based sensors like GelTip~\cite{GelTip} and Romero et al.~\cite{Romero2020GelSight2.5} solve the stability problem with a smooth and uniform support structure out of transparent glass, acrylic, or resin~\cite{Romero2020GelSight2.5}, allowing for good imaging quality and acting as a light guide. Our metal skeleton can be designed independent of the lighting.

Moreover, \method{}'s shell is hollow so that the entire system is lightweight; avoiding direct contact with any optical elements (in contrast to \cite{OmniTact}) also reduces the chances of image distortion and system damage.
Constructing a single elastomer layer that serves all purposes is a simple, compact, robust, and wireless solution for haptic sensing.
All other vision-based haptic sensors are built from multiple layers of different materials. Durability issues often arise due to non-permanent attachment between layers, \eg, \cite{GelSight,GelSlimFatigue,GelTip,OmniTact}. Another difference is that our elastomer layer is opaque enough to block all interference from ambient light, ensuring reliable output.
To demonstrate the sensitivity of our chosen approach, we include a thin, flat area of elastomer near the sensor's end for higher-resolution perception of detailed shapes (akin to a tactile fovea).
\begin{figure}
    \includegraphics[width=\textwidth]{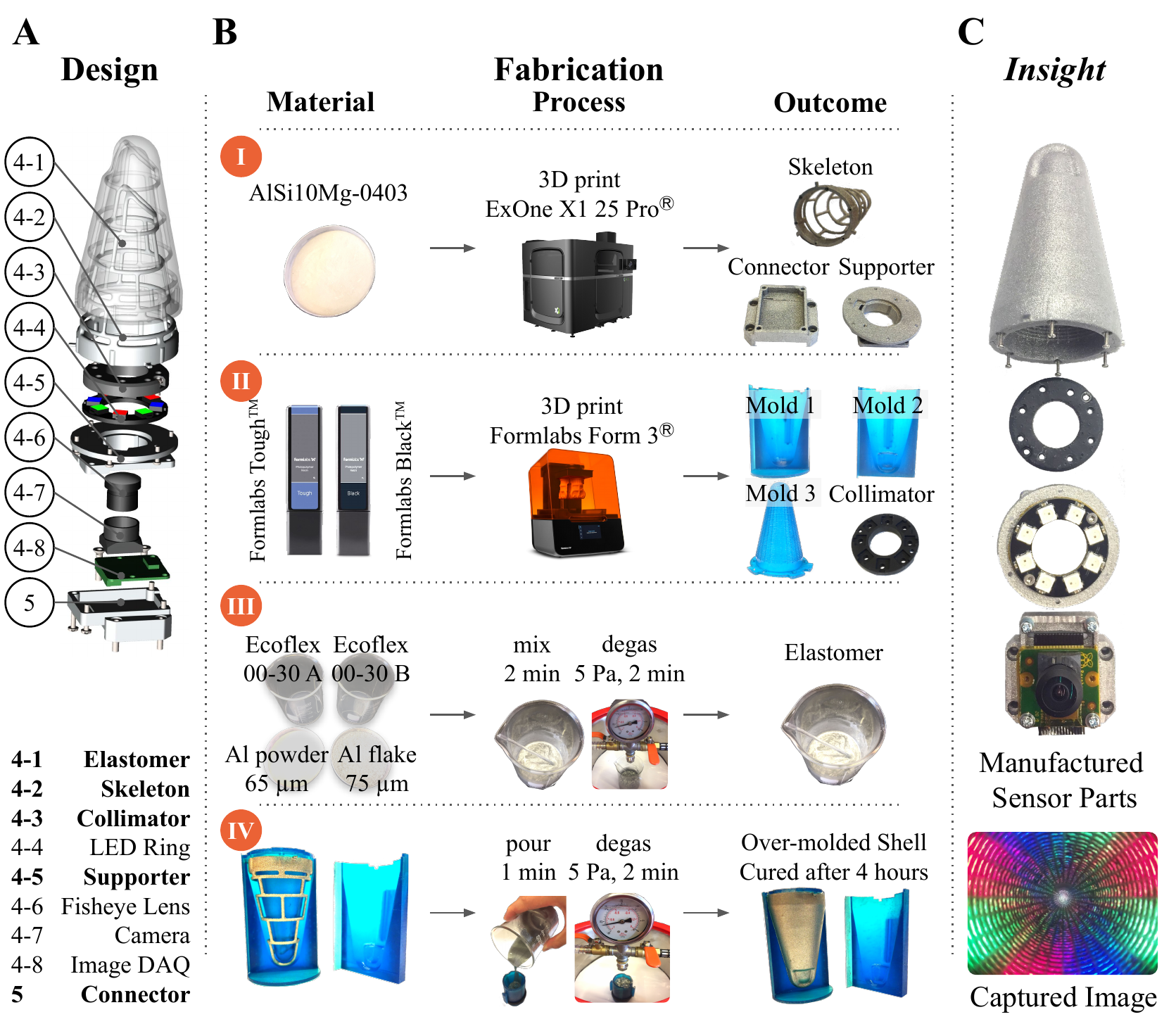}
    \caption{\textbf{The fabrication process of \method{}}.
    \textbf{A} provides an exploded view of \method{} with all parts in the design; bold items were custom-fabricated.
    \textbf{B} shows the materials, processing steps, and intermediate outcomes for the elastomer, the skeleton frame, the molds for over-molding, the collimator, and the over-molded sensing surface.
    \textbf{C} presents the partially assembled \method{} and an image captured under no contact.}
    \label{fig:Figure2}
\end{figure}

\paragraph{Imaging} Two main techniques can be used to obtain 3D information from a single camera.
Photometric stereo (PS)~\cite{PS} uses multiple images of the same scene with varying disparate light sources from different illumination directions to infer the 3D shape from shading information.
Structured light (SL)~\cite{SL} is a single-shot 3D surface-reconstruction technique that uses a unique light pattern and the fact that its appearance depends on the shape of the 3D surface on which it is projected.
Generally, PS is better at capturing local details, while SL is used for coarser global reconstruction~\cite{PSSL,PSSL2}.
PS is most effective when the illumination is nearly parallel to the surface, where the normal vectors of the deformed surface can be finely reconstructed from shading information~\cite{GelSight}.
Sensors built on PS~\cite{GelSight,GelSlimFatigue, GelTip} employ light guides to create this desirable tangential surface illumination, which is challenging for highly curved sensing surfaces, as discussed in~\cite{GelTip,Romero2020GelSight2.5}.
SL allows for more perpendicular lighting of the surface and improves with larger disparity between light source and camera.

\method{} is the first haptic sensor that combines PS and SL to detect the deformation of a full 3D cone-shaped surface in the single-camera single-image setting.
LED light sources around the camera produce distinct light cones (eight in our prototype, as shown in \figS{fig:Figure1}{C}).
The lighting direction is adjusted through a collimator to introduce a suitable SL pattern that favors locally parallel lighting for PS, as depicted in \fig{fig:sup:light-properties}{C}.
In contrast to the light guides used in other sensors~\cite{GelSight,GelSlimFatigue,GelTip,Romero2020GelSight2.5}, our collimator allows for flexible lighting and is independent of the support structure.
When an area of the sensor surface is contacted from the outside, the surface orientation changes, which causes a difference in color intensity through shading.
The surface displacement additionally changes the distance of the surface to the camera, which can be detected with SL cones due to the color change per pixel.

\paragraph{Information}
Sensors can capture many different types of haptic information, such as vibration~\cite{ContactInfo-Vibration}, deformation~\cite{HapDefX,ContactInfo-Defo,GelSight}, undirected pressure distribution~\cite{Resistive}, and directional force distribution~\cite{SlimFEM,GelForce}.
For robotics applications, a directional force distribution is the preferred form of contact information, as it describes the location and size of each contact region, as well as the local loading in the normal and shear directions~\cite{TacInfo}.
Our proposed sensor is the first sensor designed to deliver this type of contact information, \ie a 3D directional force distribution over a 3D conical sensing surface represented by a fine mesh of points, where each point has three force elements that are orthogonal to one another.

In a classical estimation chain, the force distribution is inferred from the surface displacement using a linear stiffness matrix based on elastic theory~\cite{TheoryofElastics}.
The displacement map can be acquired by analytically reconstructing the current normals of the sensing surface or numerically deriving the relative movement of labeled markers from the raw image captured by the camera, as done in~\cite{SlimFEM,GelForce}.
However, large deformations violate linearity between displacement and force.
In addition, the over-molding in our design creates an inhomogeneous surface, where the stiffness matrix is difficult to model accurately.
Shear forces are visible as small lateral deformations that highly depend on the distance to the stiff skeleton.
Moreover, the reconstruction of surface normals requires evenly distributed light, without shadows or internal reflections~\cite{GelSight}.
Tracking markers~\cite{Beads,Beads2,GelForce} rather than a surface does not solve the fundamental problems with displacement-focused approaches.

Thus, we employ a data-driven method to estimate the force distribution directly from the raw image input using machine-learning techniques, namely an adapted ResNet~\cite{ResNet}, which is a favored deep convolutional-neural-network architecture.
To collect reference data to train the neural network, we built a position-controlled 5-DoF test bed with an indenter that probes the designed sensor.
A 6-DoF force-torque sensor (ATI Mini40) measures the force vector applied to the indenter so that we can simultaneously record \emph{ground truth} forces and corresponding images from the camera inside the sensor.
The target force distribution map corresponding to each contact is computed by a simple spatial approximation using the known force vector, contact location, and indenter diameter.
The approximation was chosen from a set of five candidates based on the resulting contact inference performance, as detailed in~\sec{sec:app:approximation}.
A subset of all data is used to train the machine-learning model.
The entire process is illustrated in \figS{fig:Figure1}{D} and detailed in the Methods section.

\paragraph{Fabrication}
As depicted in~\fig{fig:Figure2}, the fabrication process of \method includes three main aspects: the imaging system, mechanical components, and optical properties.
An explanation of the design choices and further details of the fabrication process can be found in the Methods section.

\subsection{Performance}
The performance of the sensor is evaluated with respect to both accuracy and sensitivity.
The first measure of accuracy is direct single-contact estimation: a contact force needs to be localized, and its magnitude and direction must be inferred.
Second is force distribution estimation for single contact: the contact area and directional force distribution over the entire sensing surface are inferred.
In addition, we provide qualitative results for multiple contacts.
Lastly, we evaluate \method{}'s sensitivity by studying whether it can perceive gravitational effects and characterizing its ability to detect shapes contacting the high-sensitivity zone.

\begin{figure}
    \includegraphics[width=\textwidth]{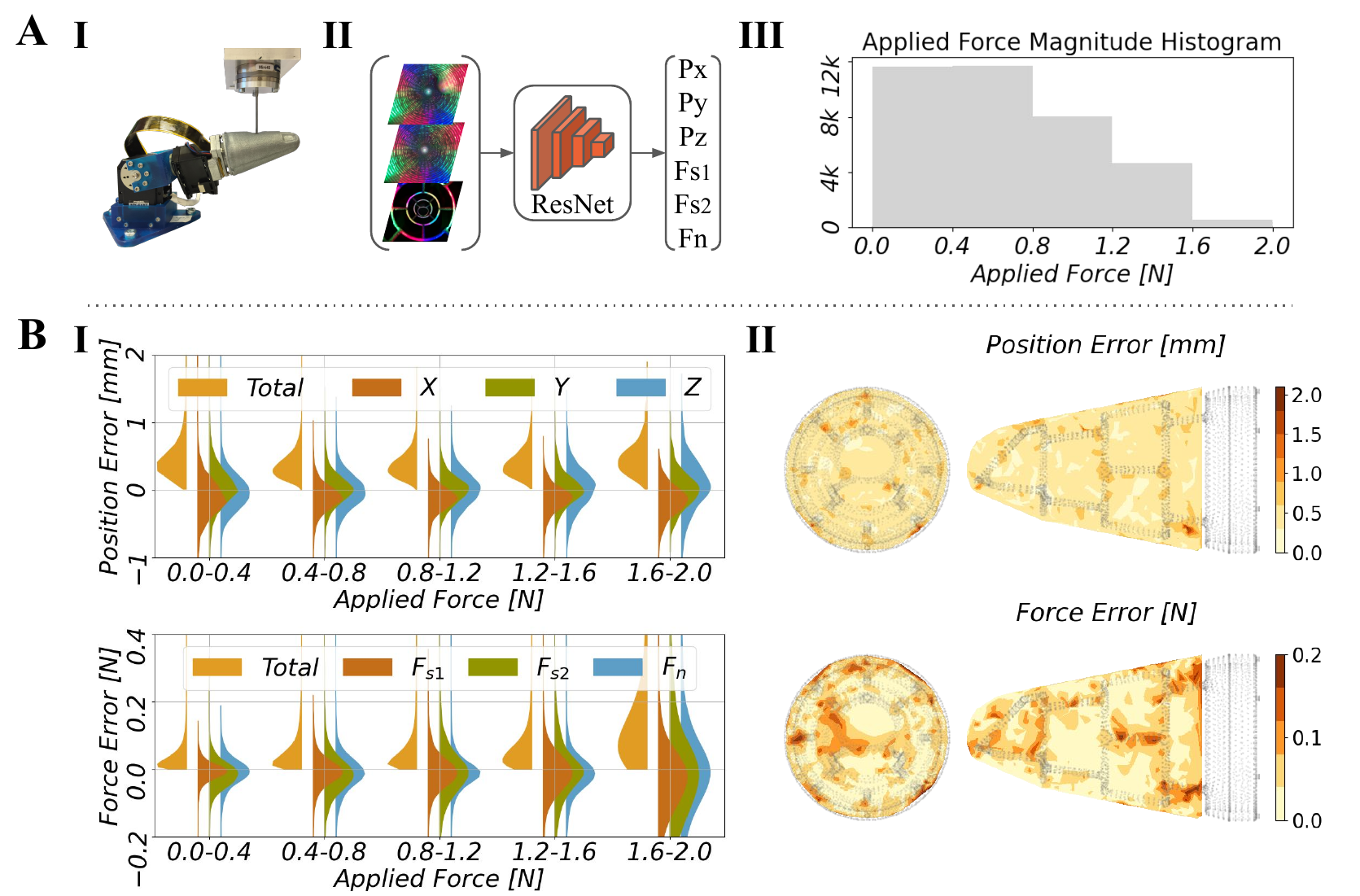}
    \caption{\textbf{Single contact performance with direct estimation}.
    \textbf{A}: the estimation pipeline for inferring single contact position and force.
    \textbf{A-I} shows the real experimental setup, in which the test bed probes \method{} and collects data.
    \textbf{A-II} sketches the machine-learning model: the inputs are three images (raw image, reference image, and skeleton image), and the outputs are the contact location and contact force vector.
    \textbf{A-III} shows a histogram of the forces applied in the data collection procedure.
    \textbf{B}: statistical evaluation of the sensor's performance on the test data.
    \textbf{B-I} presents the localization and force estimation performance grouped by applied force magnitude.
    The red-, green-, and blue-colored half-violins show the distribution of deviations in the $x$, $y$, and $z$ directions, respectively. The force is predicted relative to the surface in normal direction $F_n$ and two shear directions $F_{s1}$ and $F_{s2}$.  The orange half-violins stand for the resulting total errors.
    \textbf{B-II} indicates the spatial distribution of the localization and force quantification errors for the same test data.}
    \label{fig:Figure3}
\end{figure}

\paragraph{Accuracy: direct contact estimation}
Our primary way to measure haptic sensor accuracy is to quantitatively evaluate the system's ability to localize contacts and measure the applied force.
We employ a machine-learning-driven pipeline.
First, we use a hemispherically tipped indenter with a diameter of $4\,$mm to probe a large number of points distributed across \method{}'s sensing surface (\figS{fig:Figure3}{A-I}).
In this procedure, we collect the images under contact and the contact force vectors from the ATI Mini40 force sensor, as well as the position of each contact on the sensor's surface using our 5-DoF test bed (\figS{fig:Figure1}{D}).
The histogram of the applied forces in \figS{fig:Figure3}{A-III} shows that most contacts have magnitudes smaller than $1.6\,$N, as we set this value as the threshold of data collection to avoid damaging the sensor.
Then we train a machine-learning model (modified ResNet~\cite{ResNet} structure) to infer the contact information.
The inputs to the model are the image under contact, a static reference image without contact, and a static image of the stiff skeleton for inhomogeneous elasticity encoding (recorded before over-molding in a dark environment).
The outputs are the 3D coordinates of the contact in the sensor's reference frame and the 3D force components expressed in the local surface coordinate frame, as depicted in \figS{fig:Figure3}{A-II}.
Details about data collection and machine learning are summarized in the Methods section.\looseness-1

We evaluate the single-contact direct estimation accuracy of localization and force sensing for an applied force magnitude up to $2.0\,$N, as shown in \figS{fig:Figure3}{B-I}.
All reported numbers are for test contact points that do not appear in the training data.
The overall median localization precision is around $0.4\,$mm, and the force magnitude precision is approximately $0.03$\,N in the normal and shear directions.
The force direction is estimated with a precision of approximately $5^\circ$.
Notice that the test bed has an overall position precision of $0.2\,$mm, and the force-torque sensor has a force precision of $0.01/0.01/0.02\,$N ($F_x/F_y/F_z$).
\method{}'s accuracy in localization is remarkably stable over different force ranges,
while the error in force amplitude slightly increases with higher interaction force.
For strong applied forces (over $1.6\,$N), the force accuracy becomes worse,
presumably because we have little training data for this domain (histogram in \figS{fig:Figure3}{A-III}).
Another explanation is that high forces occur most often at locations near the stiff frame (\figS{fig:Figure3}{B-II}), which deforms only a little.
There is no noticeable difference in the localization and force accuracy in the sensor frame's $x$, $y$, and $z$ directions.

We particularly evaluate \method{}'s accuracy at localizing test contact points, as shown in \figS{fig:Figure3}{B-II}.
The accuracy is stable across the entire surface, and higher errors appear near the stiff frame.
Only areas near the camera show a systematic performance drop;
because our camera has a $4$:$3$ aspect ratio, it cannot see two opposite areas at the base of the shell, below the lowest ring of the stiff frame.

\paragraph{Accuracy: force map estimation}
To infer contact areas and multiple simultaneous contacts, we now consider the distribution of contact force vectors across the entire surface, which we call a whole-surface force map.
Altogether, the force map yields valuable information for robotic grasping and manipulation, \eg for slip detection, in-hand object movement, and haptic object recognition.

The \method{} sensor has a 3D curved surface and thus needs to output a force map with the same shape.
We create a fine mesh of points spanning the entire surface with an average distance of $1\,$mm between neighboring points.
For the results reported here, we use $3800$ points.
Each point has three output values describing the force components it feels in the $x$, $y$ and $z$ directions expressed in the reference frame of the sensor.

\begin{figure}
    \includegraphics[width=\textwidth]{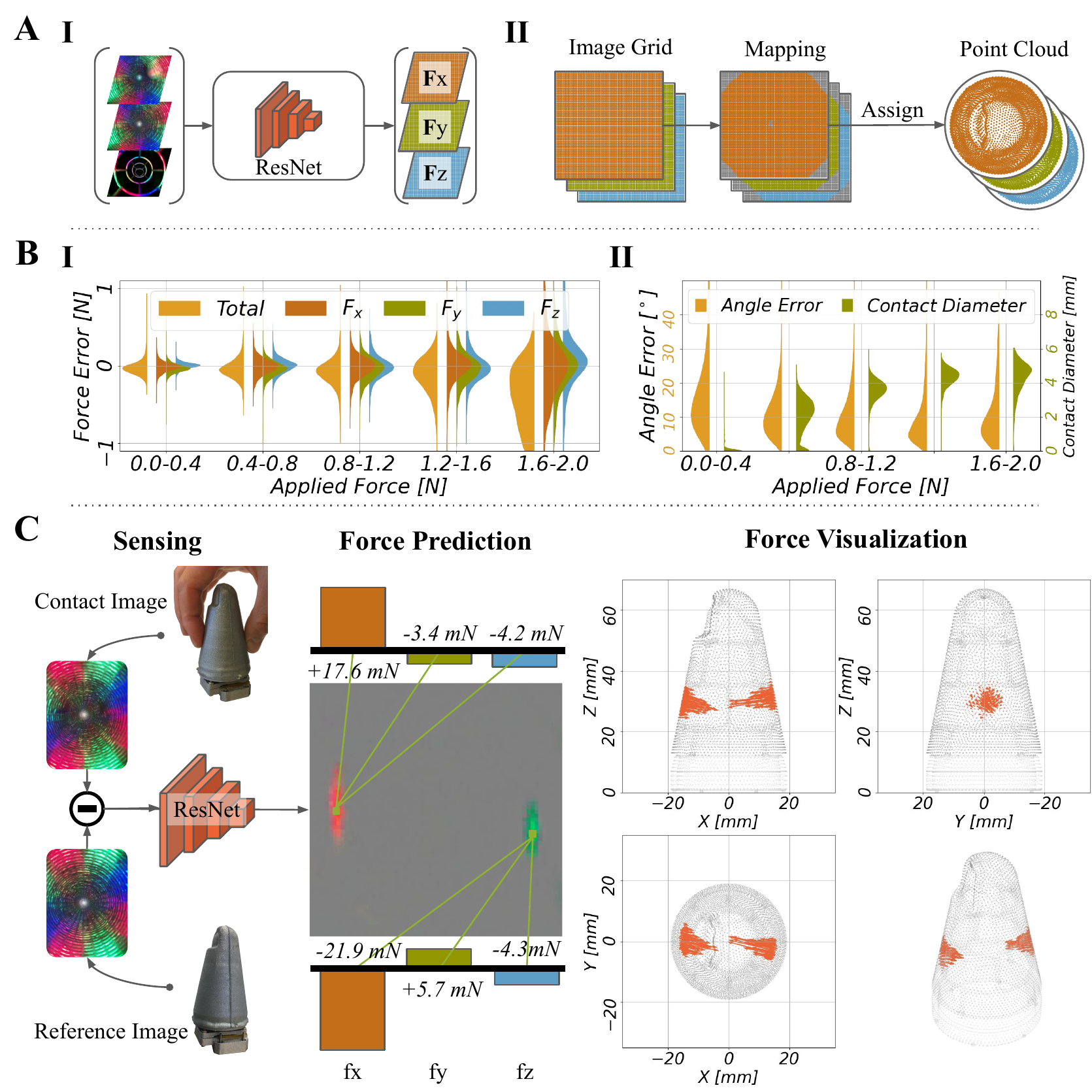}
    \caption{\textbf{Performance evaluation of the force map.}
    \textbf{A} indicates the pipeline of estimating the force distribution; the ResNet network transforms three images (raw, reference, skeleton) into the x-y-z force map image (\textbf{A-I}), and its pixels are mapped to points on the sensing surface (\textbf{A-II}).
    \textbf{B} shows the quantitative evaluation of the performance for force amplitude, force direction, and contact area size inference grouped by applied force amplitude.
    \textbf{C} demonstrates the data flow and estimated force map when the sensor is pinched and rotated by two fingers.}
    \label{fig:Figure4}
\end{figure}

Similar to the direct contact estimation, we also employ a machine-learning-driven pipeline.
Instead of the six-dimensional output~(\figS{fig:Figure3}{A-II}), the network now produces the approximate force distribution map~(\figS{fig:Figure4}{A-I}) using only convolutional layers.
The map is estimated as a flat image with three channels ($\mathbf{F_x}, \mathbf{F_y}, \mathbf{F_z}$) to describe nodal forces (individual force on each point) in the $x$, $y$, and $z$ directions, respectively, mimicking the red, green, and blue channels in a colorful image.
Each pixel in the image corresponds to one point in the force map.
The correspondence is established using the Hungarian assignment method~\cite{Hungarian}, which minimizes the overall distance between pixels and points projected to the 2D camera image, as shown in \figS{fig:Figure4}{A-II}.
Training the machine-learning model from collected data additionally requires target force distribution maps (\figS{fig:Figure1}{D}).
Since they are not measured directly, we approximate the force map applied by the indenter by distributing the measured total force locally across the surface.
From a set of five diverse candidates, the approximation yielding the best performance in localization and force magnitude accuracy is selected (see \fig{fig:sup:approximation}, \tab{tab:sup:approximation}).

The quantitative estimation accuracy for the force amplitude and force direction are reported in \figS{fig:Figure4}{B} grouped by force magnitude.
The evaluation is based on the comparison between the three-dimensional force vectors summed across the predicted force map and the ground-truth force vectors using the same single-contact data set.
The median error in inferring the total force is around $0.08\,$N, and the error grows with increasing force (\figS{fig:Figure4}{B-I} and \fig{fig:sup:accuracy}).
The system's tendency to slightly underestimate larger forces is likely caused by our force map approximation method, the influence of the skeleton, and the machine-learning method itself, which tends to estimate smooth force distributions rather than peaked maps.
An ablation of the skeleton image as input leads to worse underestimation (\supp{sec:app:ablation}), supporting part of this hypothesis.
The median error in inferring the force direction is around $10^\circ$ for low contact forces, and it decreases to $5^\circ$ with higher applied forces (\figS{fig:Figure4}{B-II}).
Moreover, we can also localize the contact with a precision around $0.6\,$mm based on the force map by averaging the locations of the $20$ points with the highest force amplitudes (\fig{fig:sup:accuracy}).
\supp{sec:app:ablation} analyzes how this performance depends on the amount of data and the type of input provided to the network.

\begin{figure}
    \vspace{-2cm}
    \includegraphics[width=\textwidth]{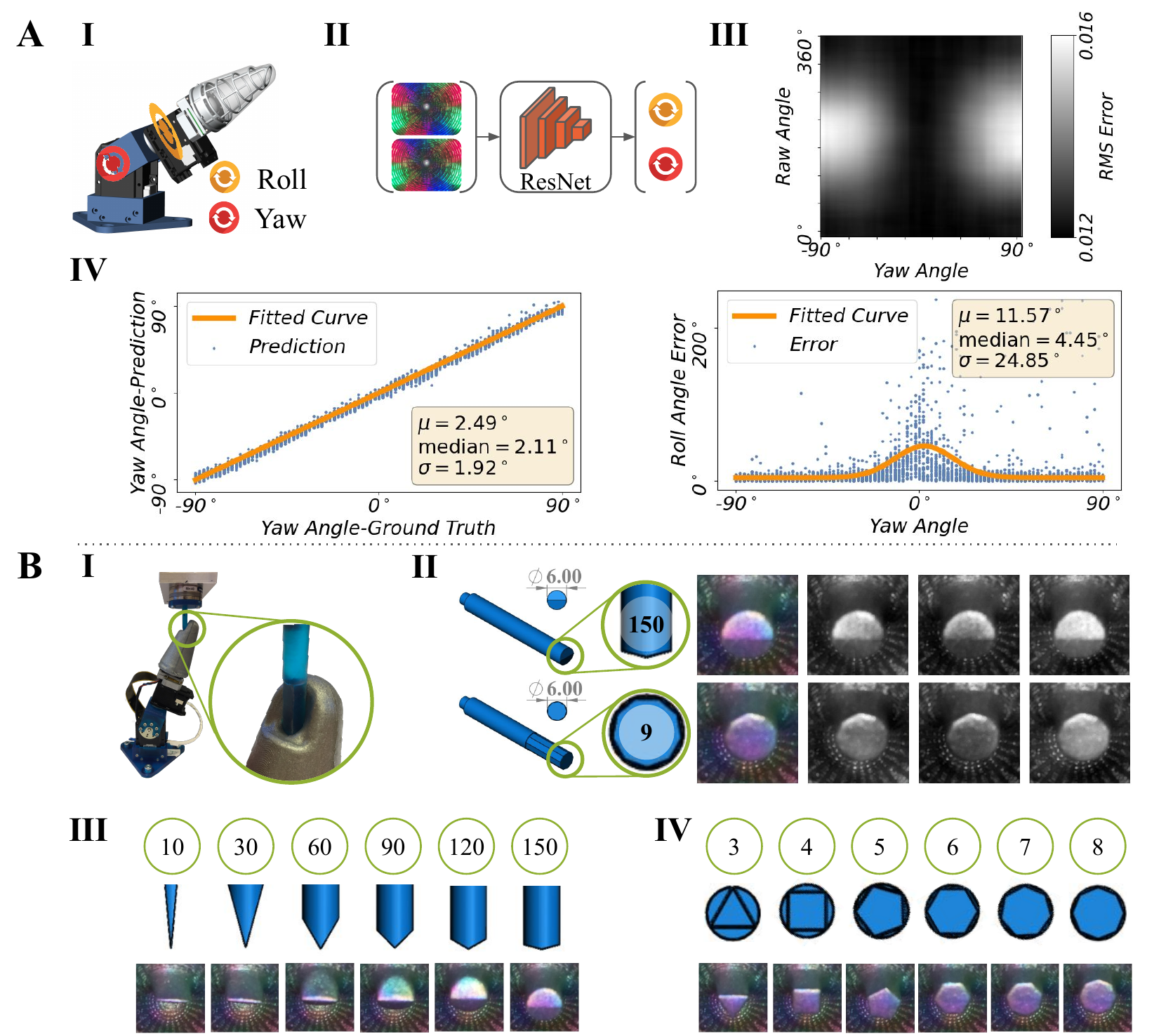}
    \caption{\textbf{Evaluation of \method{}'s sensitivity to self-posture and detailed shapes}.
    \textbf{A}: quantitative evaluation of sensor posture recognition.
    \textbf{A-I} and \textbf{A-II} present the experiment setup and the inference procedure for posture (roll and yaw angles).
    The network maps the difference between the current image and the reference image to posture coordinates.
    \textbf{A-III} shows the pixel-wise root-mean-square (RMS) of the image difference as the sensor is rotated to all possible roll and yaw angles.
    \textbf{A-IV} summarizes the posture estimation accuracy statistically: the yaw angle estimation performance with a yellow fitted curve (left) and the roll angle estimation accuracy under different yaw angles (right).
    \textbf{B}: qualitative evaluation of the tactile fovea for shape detection.
    \textbf{B-I} shows the experiment setup for applying differently shaped probes to the high-sensitivity region.
    \textbf{B-II} presents the system's perceptual limits for sharpness (v-shaped wedge with an included angle of $150^\circ$) and number of edges (nine-sided polygon) with an indenter diameter of $6\,$mm, along with their corresponding captured raw images and respective red, green, and blue channels.
    \textbf{B-III} and \textbf{B-IV} depict the sharpness and edge tests: included angle and edge count (upper), indenter samples (middle), and captured images under indentation tests (lower).}
    \label{fig:Figure5}
\end{figure}

The contact area is estimated by identifying the points with predicted forces larger than $0.02\,$N.
The diameter of this contact area increases with higher applied force and tends to overestimate by about $1\,$mm for a $4$-mm indenter at high forces.
Our \method{} prototype possesses a nail-shaped zone with a thinner elastomer layer ($1.2$\,mm) and a sensing area of $13\times11\,$mm$^2$, as indicated in \figS{fig:Figure5}{B}.
The median position and force errors in the tactile fovea are $0.3\,$mm and $0.026\,$N over an applied force range of $0.03-0.8\,$N, which shows better position accuracy and force accuracy than other sensing areas.

Furthermore, we use an indenter with $12\,$mm diameter to validate the force map inference performance and report details in \fig{fig:sup:largerindenter}.
The median position accuracy is $1\,$mm. For higher applied force, the underestimation of force magnitude is more pronounced.
Force direction is measured to a high level of accuracy, achieving a median error of $8^\circ$. The median contact area estimates closely match the size expected for a $12$\,mm indenter at each force level. As anticipated, the predicted force map is inhomogeneous and shows higher forces near the beams of the skeleton.

\paragraph{Multiple simultaneous contacts}
We also qualitatively demonstrate the sensor's performance during multiple complex contacts, as shown in \figS{fig:Figure4}{C} and \vid{S4}.
The figure shows how the captured image and a reference frame without contact are combined to yield the system's perceptual response to a human using two fingers to pinch and slightly twist the sensor.
\fig{fig:sup:multiple} and \vid{S4} show the response for many different contact situations.
Each pixel of the force map contains the three force values estimated at that point.
To facilitate interpretation, we also visualize each contact force vector on the 3D surface of the sensor.
The experimenter's counter-clockwise twisting input can be seen in the slant of the force vectors when the sensor is viewed axially.
In our experiments, the sensor was consistently able to discriminate up to five simultaneous contact points and estimate each contact area in a visually accurate manner.
\vspace*{-.5em}
\paragraph{Sensitivity} The final two experiments evaluated \method{}'s sensitivity to subtle haptic stimuli. The sensor can accurately estimate its own orientation relative to gravity by visually observing the small gravity-induced deformations of the over-molded elastomer (see \figS{fig:Figure5}{A}, \figS{fig:sup:sensitivity}{A}, and \supp{sec:app:data-interpret}). Note that this experiment was conducted without any contacts and in a dark room to rule out other possible clues about self-posture. The median error for predicting yaw was 2.11$^\circ$, and it was 4.45$^\circ$ for roll, with the highest errors for the roll angle around vertical, as expected for this problem. The camera was also found to capture relevant shape details when v-shaped wedges and extruded polygons were pressed into the tactile fovea (\figS{fig:Figure5}{B}).

\section{Discussion}
We present a soft haptic sensor named \method{} that uses vision and learning to output a directional force map over its entire thumb-shaped surface.
The sensor has a localization accuracy of $0.4\,$mm, force magnitude accuracy of $0.03\,$N, and force direction accuracy of $5^\circ$.
It can discriminate the locations, normal forces, and shear forces of multiple simultaneous contacts -- up to five regions in our evaluation.
Moreover, the sensor is so sensitive that its quasi-static orientation relative to gravity can be inferred with an accuracy around $2^\circ$.
A particularly sensitive tactile fovea with a thinner elastomer layer allows it to detect contact forces as low as $0.03\,$N and perceive the detailed shape of an object. A detailed comparison between \method{} and other sensors can be found in~\tab{tab:Overview} and the Methods section.

The majority of sensors detect deformations with classical methods and use linear elastic theory to compute interaction forces.
This approach requires good calibration and special care when it comes to reflection effects and inhomogeneous lighting.
The linear relationship between deformations and forces is often violated for strong contacts and for inhomogeneous surfaces like the over-molded shell of \method{}.
Because our method is data-driven and uses end-to-end learning, all effects are modeled automatically.
The downside of our approach is that it requires a precise test bed to collect reference data.
Once constructed, the test bed can be used to collect data for different sensor geometries -- only a geometric model of the design is required.

The inhomogeneity of our sensor's surface might cause unwanted effects in some applications.
Robotic systems that move with high angular velocities and high accelerations will likely see tactile sensing artifacts caused by inertial deformations of the soft sensing surface; data collected during dynamic trajectories can potentially mitigate these effects.

In general, our sensor design concept can be applied and extended to a wide variety of robot body parts with different shapes and precision requirements.
The machine-learning architecture, training process, and inference process are all general and can be applied to differently shaped sensors or other sensor designs.
We also provide ideas on how to adjust \method{}'s design parameters for other applications, such as the field of view of the camera, the arrangement of the light sources, and the composition of the elastomer.

\section{Methods}\label{sec:materials_methods}
We conducted several experiments to make informed design choices and validate the functionality of \method{}.

\paragraph{Sensor shape and camera view}
The sensor is cone-shaped with a rounded tip to allow all-around touch sensation in a structure similar to a human thumb.
The sensor has a base diameter of $40\,$mm and a height of $70\,$mm.
The employed camera is a Raspberry Pi camera V2.0 (MakerHawk Raspberry Pi Camera Module 8\,MP).
The camera's resolution is $1640 \times 1232$, captured at a frame rate of $40$ frames per second.
With a $160^\circ$ fisheye lens, the camera's field of view is $123.8^\circ \times 91.0^\circ$. See \fig{fig:sup:GeometryandFoV} and \tab{tab:sup:FoV} for additional details about the camera.
Multiple cameras are suggested if the whole sensing surface cannot be seen by a single camera, as done in OmniTact \cite{OmniTact}, at the cost of increased wiring, material costs, and computational load.

\paragraph{Light source and collimator}
We use a commercial LED ring that contains eight tri-color LEDs (WS2812 5050).
The colors of the eight LEDs are programmed to be red (R), green (G), blue (B), R, G, R, B, G in circumferential order, and the relative brightnesses for the R, G, and B light sources are  $1:1:0.5$, respectively.
We designed a 3D-printable collimator (3D printer: Formlabs Form 3 \textsuperscript{\textregistered}, Material: Standard Black \textsuperscript{TM}, Note: Formlabs owns the trademark and copyright of these names and pictures used in \fig{fig:Figure2}{B-II}) with a tuned diameter ($2.5\,$mm) and a radially tilted angle ($3^\circ$) toward the outside to constrain the light-emitting path and create the structured light distribution (see \fig{fig:sup:light-properties}).
Detailed analysis can be found in \sec{sec:app:light-source}.

\paragraph{Soft surface material, skeleton, and over-molding}
\method{}'s mechanical properties are optimized to ensure high sensitivity to contact forces, robustness against high impact forces, and low fatigue effects.
For the elastomer, we choose Smooth-On EcoFlex~00-30 silicone rubber as the moldable material for the soft sensing surface because it is readily available and has a high elongation ratio of $900\%$ (\tab{tab:sup:Mechanical-Elastomer}).
The skeleton is made of AlSi10Mg-0403 aluminum alloy, which can withstand forces up to $40~\,$N in the shape of our prototype structure (\tab{tab:sup:Mechanical-Skeleton}).
These two materials are chosen based on their material data sheets and finite element analysis (FEA) results~\cite{ANSYS}.
The elastomer is cast using 3D-printed molds (3D printer: Formlabs Form 3~\textsuperscript{\textregistered}, Material: Tough \textsuperscript{TM}, Note: Formlabs owns the trademark and copyright of these names and pictures used in \fig{fig:Figure2}{B-II}), and the skeleton is 3D-printed in Aluminum (3D printer: ExOne X1 25 Pro \textsuperscript{TM}, Material: AlSi10Mg-0403, Note: ExOne GmbH owns the trademark and copyright of the name and picture used in \fig{fig:Figure2}{B-I}).
We combine the skeleton and the elastomer without adhesive by over-molding, as described in \figS{fig:Figure1}{B} and \figS{fig:Figure2}{B}.
Because of the working principle, there is no special treatment required for the molds; they are used straight from the 3D printer, in contrast to, \eg, Romero et al.~\cite{Romero2020GelSight2.5}. Furthermore, the manufacturing procedure is simple and requires only a single step.

The diameter of the skeleton beams and the thickness of the surrounding elastomer are optimized for robustness, as described in \fig{fig:sup:mechanical}.
FEA revealed that we can improve the system's sensitivity to contact forces by positioning the skeleton not in the center of the elastomer layer but closer to the inner surface.

\paragraph{Optical properties}
We need a material with the right reflective properties (surface albedo, specularity) for the sensing surface.
On one hand, it should not be too reflective because reflections saturate the camera and diminish sensitivity.
On the other hand, no point on the surface should be very dark, because the camera needs to detect changes in reflected light.
Moreover, the material has to prevent ambient light from perturbing the image and deteriorating the sensing quality.
The sensing surface is made of a flexible and moldable translucent elastomer mixed with aluminum powder and aluminum flakes.
The aluminum powder makes the surface opaque to ambient light, and the aluminum flakes adjust the reflective properties, as shown in \figS{fig:Figure1}{C}, \figS{fig:Figure2}{B}, \fig{fig:sup:material-mixture} and \tab{tab:sup:Coating}.
Aluminum powder with $65\,\mu$m particle diameter and aluminum flake with $75\,\mu$m particle diameter are mixed with the EcoFlex~00-30 in a weight ratio of $20:3:400$ to ensure proper light reflection properties for \method (see \fig{fig:sup:material-mixture} and \tab{tab:sup:Coating}).
Note that the resulting mixture is opaque, so that the sensor's inside is fully shielded from ambient light.

\paragraph{Finite element analysis}
To analyze the over-molded shell's mechanical properties during the design process, we built a suitable finite-element model using Ansys~\cite{ANSYS}.
The model includes suitable values for Young's modulus and Poisson's ratio for both EcoFlex\,00-30 ($70\,$kPa, $0.4999$) and the aluminum alloy ($75\,$GPa, $0.33$)~\cite{FEM-value}.

\paragraph{Test bed}
We created a custom test bed with five degrees of freedom (DoF); three DoF control the Cartesian movement of the probe ($\vec{x}, \vec{y}, \vec{z}$) using linear guide rails (Barch Motion) with a precision of $0.05\,$mm, and two DoF set the orientation of the sensor (yaw, roll) using Dynamixel MX-64AT and MX-28AT servo motors with a rotational precision of $0.09^\circ$, which results in a translational precision of $0.2\,$mm at the tip of the sensor.
The probe is fabricated from an aluminum alloy and is rigidly attached to the Cartesian gantry via an ATI Mini40 force/torque sensor with a force precision of $0.01/0.01/0.02\,$N ($F_x/F_y/F_z$).
\method{} is held at the desired orientation, and the indenter is used to contact it at the desired location.

\paragraph{Data}
Measurements are collected using our automated test bed to probe \method in different locations.
To obtain a variety of normal and shear forces, the indenter is moved to a specified location, touches the outer surface, and deforms it increasingly by moving normal to the surface with fixed steps of 0.2\,mm. For each such indentation level, the indenter also moves sideways to apply shear forces (normal/shear movement ratio 2:1).
After a pause of 2 seconds to allow transients to dissipate, we simultaneously record the contact location, the indenter contact force vector from the test bed's force sensor, and the camera image from inside \method{}.
When the measured total force exceeds $1.6\,$ N, the data collection procedure at this specified location terminates and restarts at another location.
The contact location and measured force vector are combined to create the true force distribution map using the method described in \sec{sec:app:data-interpret}.
Images from \method are captured using a Raspberry Pi 4 Model B with $2\,$GB RAM.
All the data are collected and combined using a standard laptop.

\paragraph{Machine learning}
A ResNet~\cite{ResNet} structure is used as our machine-learning model.
The data for single contact includes a total of $187\,358$ samples at $3\,800$ randomly selected initial contact locations. The data set is split into training, validation, and test subsets with a ratio of $3:1:1$ according to the locations.
The data for posture estimation from gravity contains $16\,000$ measurements and is split in the same way.
We use four blocks of ResNet to estimate the contact position and amplitude directly (\fig{fig:Figure3}), two blocks to estimate the force distribution map (\fig{fig:Figure4}), and four blocks to estimate the sensor posture (\fig{fig:Figure5}).
The machine-learning models are all trained with a batch size of $64$ for $32$ epochs, using Adam with a learning rate of $0.001$ for mean squared loss minimization.
\sec{sec:app:ml-details} provides more details about the structure of the machine-learning models that we use.
The performance of the models with less training data is studied in \supp{sec:app:ablation}.

\paragraph{Operating speed}
The current version of \method{} is not optimized for processing speed. Images are captured using a Raspberry Pi 4 (Model B with 2 GB RAM) with a Python script and are transmitted to a host computer (with a GeForce RTX 2080Ti GPU) via Gigabit Ethernet. Images have a size of $1640\times1232$ and are effectively transferred at $11$\,fps and downsized to $410\times308$ using a Python script. The image processing with the deep network for the force-map prediction runs at $10$\,fps in real time. We see multiple ways to increase the operating speed, ranging from optimized code to hardware improvements (e.g., using an Intel Neural Compute Stick) to choosing a faster deep network.

\paragraph{Comparison to state-of-the-art sensors}~\label{sec:comparison}
How does \method compare to other vision-based haptic sensors?
\tab{tab:Overview} lists its performance along with that of thirteen selected state-of-the-art sensors; we first give an overview and then compare the designs.
One of the earliest vision-based sensors is GelSight~\cite{GelSight}, which has a thin reflective coating on top of a transparent elastomer layer supported by a flat acrylic plate.
Lighting parallel to the surface allows tiny deformations to be detected using photometric stereo techniques.
Further developments of this approach increased its robustness (GelSlim~\cite{SlimFEM}), achieved curved sensing surfaces with one camera (GelTip~\cite{GelTip}) and with five cameras (OmniTact~\cite{OmniTact}), and included markers to obtain shear force information~\cite{SlimFEM}.
A different technique based on tracking of small beads inside a transparent elastomer is used by GelForce~\cite{GelForce} and the Sferrazza and D'Andrea sensor~\cite{Beads,Beads2} to estimate normal and shear force maps.
ChromaForce (not listed in the table) uses subtractive color mixing to extract similar data from deformable optical markers in a transparent layer~\cite{ChromaForce}.
The TacTip~\cite{TacTip} sensor family uses a hollow structure with a soft shell, and it detects deformations on the inside of that shell by visually tracking markers.
Muscularis~\cite{Muscularis} and TacLink~\cite{TacLink} extend this method to larger surfaces, such as robotic links, by using a pressurized chamber to maintain the shape of the outer shell; they are not listed in the table because they target a different application domain.

In terms of shape recognition and level of detail, the GelSight approach provides unparalleled performance.
The tracking-based methods, such as GelForce and TacTip, are naturally limited by their marker density and thicker outer layer.
\method uses shading effects to achieve a much higher information density than is possible with markers, but its accuracy is also somewhat limited by measuring at the inside of a soft shell with non-negligible thickness.
Beyond accurately sensing contacts, the robustness of haptic sensors is of prime importance.
Without additional protection, GelSight-based sensors are comparably fragile due to their thin reflective outer coating, which can easily be damaged.
Adding another layer increases robustness, but imaging artifacts were reported to appear after about 1500 contact trials because of wear effects~\cite{GelSlimFatigue}.
We tested \method{} for more than 400\,000 interactions without noticeable damage or change in performance.

Each sensing technology imposes different restrictions on the surface geometry of the sensor.
Vision-based tactile sensors need the measurement surface to be visible from the inside, so there is typically no space available for other items inside the sensor.
The type of visual processing also matters. TacTip's need to track individual markers requires a more perpendicular view of the surface than shading-based approaches (GelSight and \method).
Soft materials deform well during gentle and moderate contact, but they do not withstand high forces if not adequately supported. GelSight uses a transparent rigid structure for support, which can lead to reflection artifacts when adapted to a curved sensing surface~\cite{GelTip}.
An alternative is high internal pressure~\cite{TacLink}, but then the observed deformations are non-local.
The over-molded stiff skeleton in \method maintains locality of deformations and withstands high forces.

To facilitate widespread adoption, tactile sensors need to be easy to produce from inexpensive components. Imaging components are remarkably cheap these days, making vision-based sensors competitive.
However, GelSight needs a reproducible surface coating and permanent bonding between all layers, which are tricky to implement correctly~\cite{GelSight,ImprovedGelSight,MultiGelsight}.
TacTip needs well-placed markers or a multi-material surface that can be 3D-printed only by specialized machines.
\method uses one homogeneous elastomer that requires only a single-step molding procedure on top of the stiff 3D-printed skeleton.
Being able to replace the sensing surface in a modular way increases system longevity; such replacement is supported by GelSight and TacTip in principle, and it is designed to be easy in \method, though we did not evaluate the quality of the results that can be obtained without retraining.

\section*{Acknowledgments}
The authors thank the China Scholarship Council (CSC) and the International Max Planck Research School for Intelligent Systems (IMPRS-IS) for supporting H.S.
G.M.\ is member of the Machine Learning Cluster of Excellence, funded by the Deutsche Forschungsgemeinschaft (DFG, German Research Foundation) under Germany’s Excellence Strategy –
EXC number 2064/1 – Project number 390727645.
We acknowledge the support from the German Federal Ministry of Education and Research (BMBF) through the Tübingen AI Center (FKZ: 01IS18039B).\\

\textbf{Author contributions:}
H.S., K.J.K. and G.M. conceived the method and the experiments, drafted the manuscript and revised it.
H.S. designed and constructed the hardware, developed fabrication methods, designed and conducted experiments, collected and analyzed the data.
G.M. and K.J.K. supervised the data analysis.
We thank Felix Grimminger and Bernard Javot for their support during the mechanical manufacturing procedure.

\textbf{Competing interests:}
H.S., G.M. and K.J.K. are listed as inventors on two PCT provisional patent applications (PCT/EP2021/050230, PCT/EP2021/050231) that cover the fundamental principles and designs of \method{}.

\textbf{Data and code availability:}
The data and code that support the findings of this study are available at \url{https://dx.doi.org/10.17617/3.6c}.

\nocite{wuethrich2020:TriFinger,HertzTheory}
\bibliographystyle{IEEE}
\bibliography{scibib}
\newpage
\begin{center}
    \textbf{\Large Supplementary Materials for\\
A soft thumb-sized vision-based sensor with\\ accurate all-round force perception}
\end{center}

\renewcommand{\thetable}{S\arabic{table}}
\renewcommand{\thefigure}{S\arabic{figure}}
\renewcommand{\theequation}{S\arabic{equation}}
\setcounter{table}{0}
\setcounter{figure}{0}
\setcounter{equation}{0}

\renewcommand{\thepage}{\arabic{page}}

\maketitle
\appendix
\paragraph{Content of the supplementary:}
\newcommand{\frefitem}[1]{\item[Page \pageref{#1}:] \fig{#1}}
\newcommand{\trefitem}[1]{\item[Page \pageref{#1}:] \tab{#1}}
\newcommand{\prefitem}[1]{\item[Page \pageref{#1}:]}

\begin{description}[itemsep=-.1em,labelwidth=5em]
    \prefitem{sec:sub:imaging:system} \sec{sec:sub:imaging:system}: Imaging system design
    \frefitem{fig:sup:GeometryandFoV} Sensor geometry and camera FoV.
    \trefitem{tab:sup:FoV}. FoV of the camera with a fisheye lens in different operating modes.
    \frefitem{fig:sup:light-properties} Analysis of the lighting system.
    \frefitem{fig:sup:material-mixture} Soft material composition.
    \trefitem{tab:sup:Coating}. Different materials for sensing surface coating.
    \prefitem{sec:sub:mechanical} \sec{sec:sub:mechanical}: Mechanical tests for the sensor design
    \frefitem{fig:sup:mechanical} Mechanical aspects of the sensor design.
    \trefitem{tab:sup:Mechanical-Elastomer}. Mechanical properties of different sensing surface material candidates.
    \trefitem{tab:sup:Mechanical-Skeleton}. Mechanical properties of different sensor skeleton material candidates.
    \prefitem{sec:app:data-interpret} \sec{sec:app:data-interpret}: Data interpretation

    \frefitem{fig:sup:morphology} Effect of the surface structure on the sensor performance.
    \frefitem{fig:sup:approximation} Force distribution approximation and evaluation.
    \trefitem{tab:sup:approximation} Evaluation of force distribution approximation.
    \frefitem{fig:sup:accuracy} Force map evaluation.
    \frefitem{fig:sup:largerindenter} Force map evaluation with a larger indenter.
    \prefitem{sec:app:ml-details} \sec{sec:app:ml-details}:  Machine Learning Details
    \prefitem{sec:app:function} \sec{sec:app:function}: Functionality illustration
    \frefitem{fig:sup:multiple} Multiple contacts examples.

    \frefitem{fig:sup:sensitivity} Sensitivity evaluation.
    \frefitem{fig:sup:slide motion} Localizing the indenter in sliding motion.
    \prefitem{sec:app:ablation} \sec{sec:app:ablation}: Ablation Studies

    \frefitem{fig:sup:ablation} Ablation studies on dataset size and network input.
\end{description}


\newpage
\section{Supplementary Text}
\label{sec:sub:main}
\subsection{Imaging system design}\label{sec:sub:imaging:system}
\begin{figure}[p]
    \includegraphics[width=\textwidth]{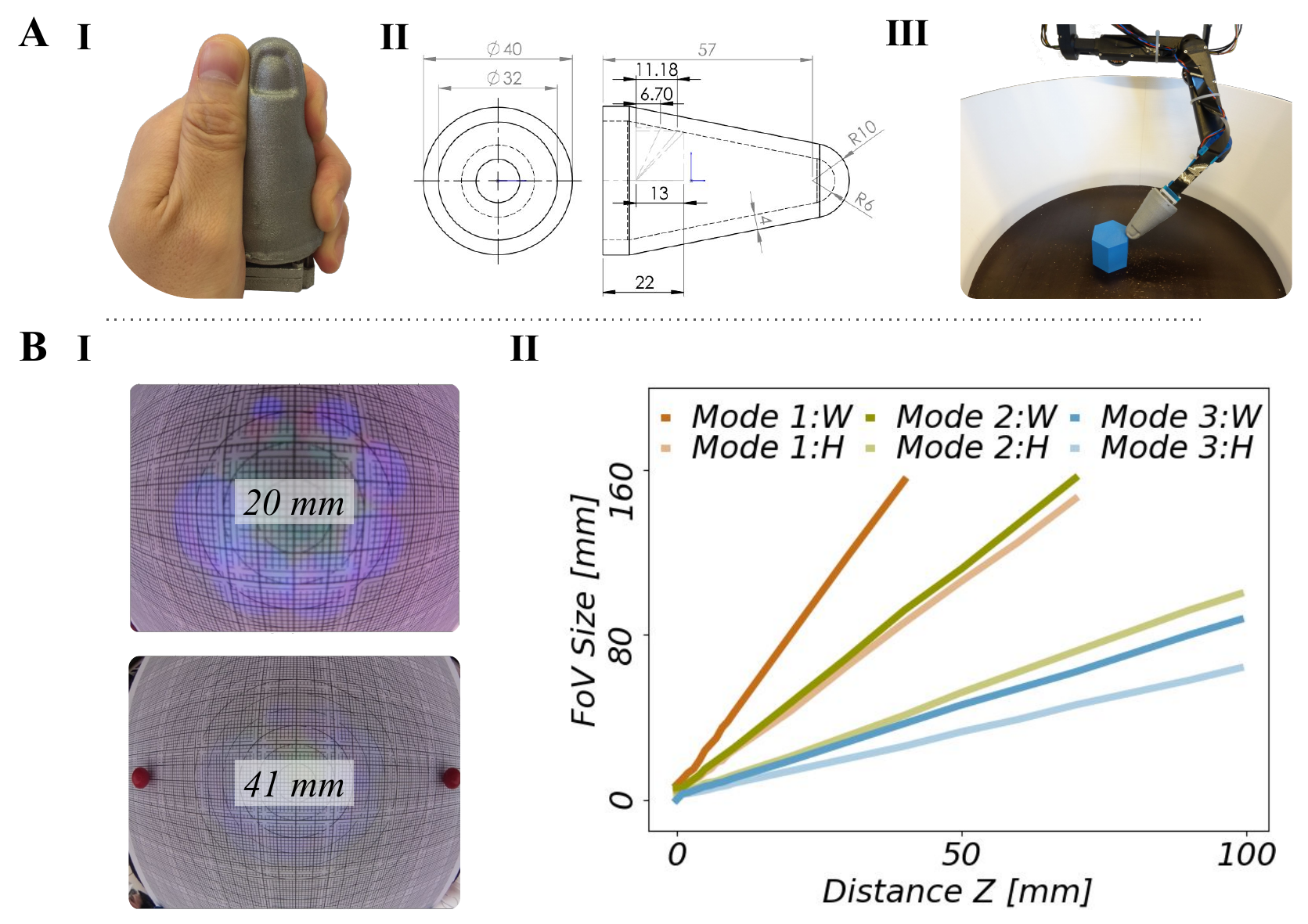}
    \caption{\textbf{Sensor geometry and camera FoV}.
    \textbf{A} shows how the sensor shape and size compare to a human thumb (\textbf{A-I}), the sensor's detailed geometry (\textbf{A-II}), and one of its intended application scenarios (\textbf{A-III}); \method{} is designed to provide haptic sensing at the tip of the depicted robot arm to facilitate dexterous manipulation.
    \textbf{B} summarizes the FoV of the camera with a fisheye lens in different operating modes.
    \textbf{B-I} presents views of a measurement grid at two distances from the camera.
    \textbf{B-II} shows the width and height of the camera's FoV in the three operating modes presented in \tab{tab:sup:FoV}.}
    \label{fig:sup:GeometryandFoV}
\end{figure}

\paragraph{Sensor geometry and FoV of the camera}
The sensor is designed with robotic manipulation platforms in mind, such as the
TriFinger manipulator~\cite{wuethrich2020:TriFinger} illustrated in \figS{fig:sup:GeometryandFoV}{A-III}.
The geometry (\figS{fig:sup:GeometryandFoV}{A-II}) is a cone shape with a maximal diameter of $40\,$mm and a height of $70\,$mm, similar to a human thumb (\figS{fig:sup:GeometryandFoV}{A-I}).
This design successfully achieves distributed 3D haptic sensing within the camera's field of view (FoV).
The sensor can be adapted to other applications by changing its size, shape, and electronics.
The following sections describe the requirements and design decisions regarding the imaging system.

To offer all-around sensing over \method{}'s 3D curved surface, the internal camera needs to see as much of the inner surface as possible.
Thus, we mount a fisheye lens with a wide-angle FoV ($160^\circ$).
The camera can operate in different modes with different frame rates (\tab{tab:sup:FoV}).
We have chosen to operate the camera in \emph{Mode 1}, mainly to maintain maximal FoV (\figS{fig:sup:GeometryandFoV}{B-I}).
However, the camera does not have equal viewing capabilities in the horizontal and vertical directions because its imaging sensor is rectangular, as shown in \figS{fig:sup:GeometryandFoV}{B-II} and \tab{tab:sup:FoV}.

\begin{table}[p]
\centering
\caption{FoV of the camera with a fisheye lens in different operating modes.}
\begin{tabular}[t]{ccccc}
\toprule
Mode & Image Resolution [W $\times$ H $\times$ fps] & FoV in W [$^\circ$]& FoV in H [$^\circ$] \\
\midrule
 \textbf{1} & \textbf{1640 $\times$ 1232 $\times$ 40}  & \textbf{123.8} & \textbf{91.0} \\
 2 & 1280 $\times$ 720 $\times$ 90 & 94.7 & 52.7 \\
 3 & 640 $\times$ 480 $\times$ 90 & 46.8 & 34.8 \\
\bottomrule
\end{tabular}
\label{tab:sup:FoV}
\end{table}%

\paragraph{Lighting system}\label{sec:app:light-source}

\begin{figure}[p]
    \vspace{-3cm}
    \includegraphics[width=\textwidth]{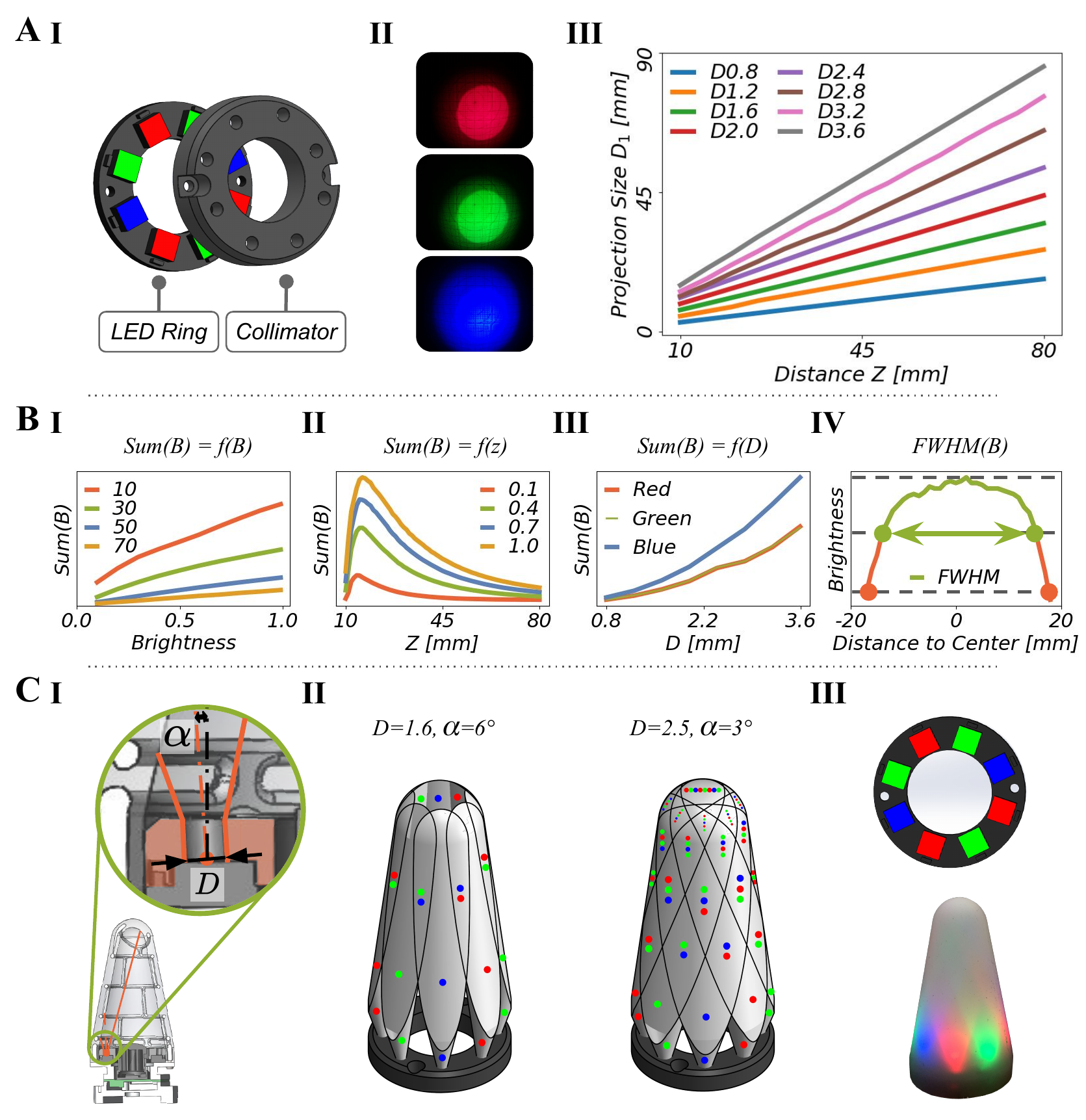}
    \caption{\textbf{Analysis of the lighting system}.
    \textbf{A} shows the correlation between collimator diameter $D$ and light cone size.
    \textbf{A-I} and \textbf{A-II} depict the LED ring, the collimator, and light patterns projected on a flat plane from different red, green, and blue light sources.
    \textbf{A-III} shows the linear scaling of the cone size.
    $Z$ is the imaging distance, $D_1$ is the diameter of the projection pattern, and different lines ($D$) are for different collimator diameters.
    \textbf{B} summarizes the light attenuation behavior.
    \textbf{B-I} shows the sum of light as seen by the camera
     depending on the surface distances ($Z$) for different brightness values of a single light source.
     \textbf{B-II} shows the same quantity as a function of the imaging distance for different brightness values.
    \textbf{B-III} shows the camera sensitivity to light brightness of different colors (red, green, blue) with varying $D$.
    \textbf{B-IV} is the light attenuation curve for a single bright disc (as in \textbf{A-II}). We use full width at half maximum (FWHM) to the quantify the size of the disc.
    \textbf{C} illustrates the effect of collimator hole size $D$ and angle $\alpha$ on the light cones and the overall light pattern.
    \textbf{C-I} depicts the details of the collimator hole geometry.
    \textbf{C-II} shows the light covering the surface area with different $D$ and $\alpha$.
    \textbf{C-III} shows the light color arrangement (R, G, B, R, G, R, B, G) and visibility in a translucent shell.}
    \label{fig:sup:light-properties}
\end{figure}
To construct the structured light pattern, we analyze the color and brightness of each light source, the behaviors of light intensity attenuation, and the parameters of the collimator as well as the camera's sensitivity to differently colored light brightness (\fig{fig:sup:light-properties}).
The light sources are generated from an LED ring with eight units, each of which has three programmable channels to create red, green, and blue light.
These units emit light in all directions of the half 3D space and light up the near field without visible differences.
We use a collimator to constrain the emitting path and construct a particular light cone for each LED, as shown in \figS{fig:sup:light-properties}{A-I} and \figS{fig:sup:light-properties}{C}.
The collimator is optimized to create a lighting pattern where most of the surface is lit by at least two and at most four LEDs.
This design enables excellent detection of deformations from the shading effects, and we avoid both over-saturated and under-lit areas.

The collimator is made from an opaque material where the holes have two key geometric parameters; see \figS{fig:sup:light-properties}{C-I}.
First, the diameter of the collimator $D$ constrains the light cone size (\figS{fig:sup:light-properties}{A-II \& III}).
Second, the tilt angle $\alpha$ slants the light cone radially outward, as shown in \fig{fig:sup:light-properties}{C-I \& II}.
We measure the effect of the collimator diameter on the light cone size with a test setup.
We use a test bed to move a marker board to defined distances $Z$ from the camera and change the collimator size.
The marker board has a fine grid on it, which is used to count the projected light cone diameter $D_1$ (\figS{fig:sup:light-properties}{A-II}). The value of $D_1$ is computed using
\emph{full width at half maximum} (FWHM), as shown in \figS{fig:sup:light-properties}{B-IV}.
The projected light cone diameter scales linearly with the collimator diameter, as expected and shown in \figS{fig:sup:light-properties}{A-III}.
Based on these curves, we test the collimator angle effect.
We tune the angle $\alpha$ ($3^\circ$) and diameter $D$ ($2.5\,$mm) jointly to make a light pattern that fully covers the internal sensing surface of \method as stated above (\figS{fig:sup:light-properties}{C-II}).

We also conduct a detailed analysis of the light attenuation behaviors; see \figS{fig:sup:light-properties}{B-I--III}.
The total received light intensity attenuates approximately linearly as the brightness reduces, and approximately quadratically as the distance increases; these effects are shown in \fig{fig:sup:light-properties}{B-I} and \figS{fig:sup:light-properties}{B-II}.
With increased distance, the light cone gets wider, but the portion of the reflected light beams seen by the camera gets smaller.
The light also attenuates within one horizontal cross-section of the light cone (\fig{fig:sup:light-properties}{B-IV}).
We use the FWHM criterion to calculate the size of the projected light cones.

Moreover, we find that the selected camera has different sensitivities to differently colored light sources: we observe a sensitivity ratio of $1:1:2$ for red, green, and blue, respectively (\figS{fig:sup:light-properties}{B-III}).
Based on this sensitivity analysis, we arranged the light sources as R, G, B, R, G, R, B, G in series.
In order to create the desired structured-light illumination, we adjust the light projection cones by tuning the collimator diameters and tilt angles, as seen in \fig{fig:sup:light-properties}{C}.

\paragraph{Sensing surface composition}
We adjust the imaging object, namely the internal surface of \method{}, based on the analysis of the light sources and camera settings.
We test the material compositions and coloring for the sensing surface by mixing aluminum powder, aluminum flakes, and pigments into the elastomer.

\begin{figure}
    \includegraphics[width=\textwidth]{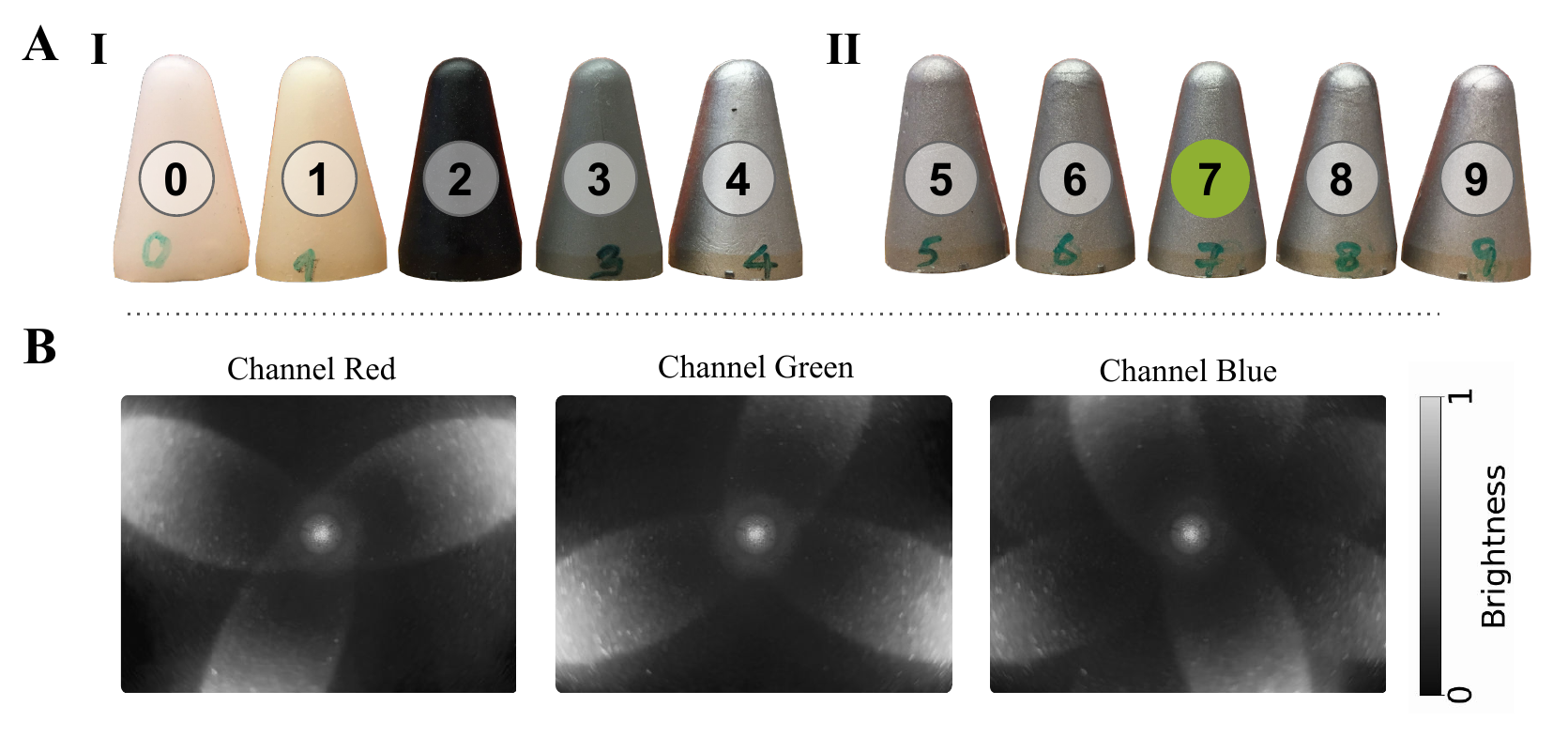}
    \caption{\textbf{Soft material composition}.
    \textbf{A} shows the ten tested versions of the EcoFlex material containing different pigments and additives, as defined in \tab{tab:sup:Coating}.
    \textbf{A-I} shows samples with no additive or only a single additive, while \textbf{A-II} presents a range of blends between aluminum powder and aluminum flake.
    \textbf{B} shows the red, green, blue channels of the light projection pattern formed by the chosen light sources shining on the inside of the selected soft shell (version 7).}
    \label{fig:sup:material-mixture}
\end{figure}

\begin{table}
\centering
\caption{Different materials for the sensing surface: ten different material compositions were considered for the coating the skeleton to create the soft sensing shell. The numbers are in given in grams [g] and correspond to the amount we used for molding one shell. Only about 25\,g of the mixture is actually needed to fill the mold.}
\begin{tabular}{ccccccccccc}
\toprule
Material & 0&1&2&3&4&5&6&\textbf{7}&8&9 \\
\midrule
 EcoFlex0030-A      &20 &20 &20  &20 &20  &20  &20  &\textbf{20}  &20  &20  \\
 EcoFlex0030-B      &20 &20 &20  &20 &20  &20  &20  &\textbf{20}  &20  &20  \\
 Aluminum Powder 65 $\mu$m  & 0 & 0 & 0  & 3 & 0  &2.0 &2.0 &\textbf{2.0} &2.0 &2.0 \\
 Aluminum Flake 75 $\mu$m   & 0 & 0 & 0  & 0 &1.5 &0.1 &0.2 &\textbf{0.3} &0.5 &1.0 \\
 Black Pigment      & 0 & 0 &0.8 & 0 & 0  & 0  & 0  &\textbf{0}   & 0  & 0  \\
 Vacuum chamber (degas)& -- & \checkmark &\checkmark &\checkmark &\checkmark &\checkmark &\checkmark &\checkmark &\checkmark &\checkmark\\
\bottomrule
\end{tabular}
\label{tab:sup:Coating}
\end{table}%
The goal is to obtain optimal reflective properties such that no part of the surface appears too dark or too bright, as described in \fig{fig:sup:material-mixture} and \tab{tab:sup:Coating}.
In \figS{fig:sup:material-mixture}{A-I}, we investigate the effect of de-gassing ($\# 0$ using de-gassing vs.~$\# 1$ without) during elastomer molding and find that degassing gives a clearer and stronger elastomer.
We compare the effect of three additives on the resulting surface's albedo: black pigment ($\# 2$), aluminum powder ($\# 3$), and aluminum flakes ($\# 4$).
We find that black pigment ($\# 2$) absorbs almost all light, so the camera can hardly see anything.
The aluminum powder ($\# 3$) shows adequate performance but gives relatively dark images.
In comparison, aluminum flakes ($\# 4$) tends to create saturated points very easily because of strong specularity.
Thus we trade-off these effects and mix aluminum powder and aluminum flake with different ratios in the elastomer, as shown in \figS{fig:sup:material-mixture}{A-II}.
We tested the light intensities for different units on the LED ring and choose composition $\# 7$ for \method{}.
\FigS{fig:sup:material-mixture}{B} shows the red, green, and blue channels for the image captured for the whole imaging system.
The red and green channels separate quite well, while the blue channel contains a bit of both red and green channel values.
This channel mixing might be due to the camera's white balance function (although it was switched off) and the elastomer's material properties.

\subsection{Mechanical tests for the sensor design}\label{sec:sub:mechanical}
\begin{figure}[p]
    \vspace{-2cm}
    \includegraphics[width=\textwidth]{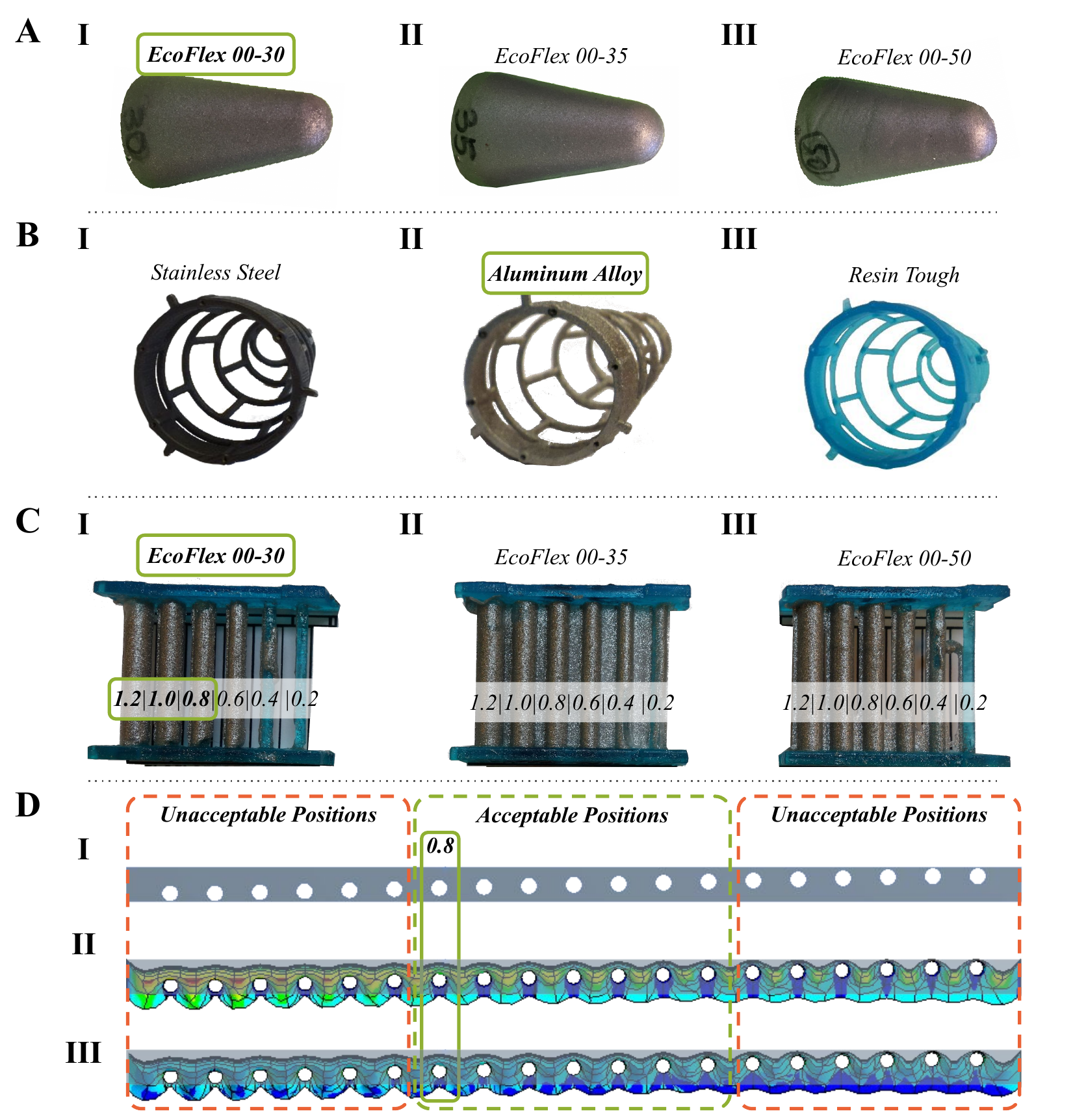}
    \caption{\textbf{Mechanical aspects of the sensor design}.
    \textbf{A} shows the candidate materials for the soft sensing shell: we choose EcoFlex~00-30.
    \textbf{B} shows the candidate materials for the stiff skeleton: we choose aluminum alloy.
    \textbf{C} presents the thickness test for the over-molding technique, which is used to connect the elastomer and the skeleton without adhesive.
    The minimal thickness for a robust connection is $0.8\,$mm.
    \textbf{D} shows a finite element analysis on how the relative position of the skeleton in the elastomer will affect the sensitivity, \ie how much deformation occurs.
    \textbf{D-I} is a soft plate containing stiff rods at varying distances to the upper and lower surfaces.
    The left and right edges and the rods are fixed.
    Homogeneous pressure is applied to the upper surface.
    \textbf{D-II} shows the resulting deformation and \textbf{D-III} the induced von Mises stress.}
    \label{fig:sup:mechanical}
\end{figure}

\paragraph{Material of the soft shell}
We choose materials from the SmoothOn EcoFlex series for the sensor elastomer due to their wide application in soft robotics and their favorable properties in terms of weight, durability and elongation ratio.
We compare three materials out of this series in \figS{fig:sup:mechanical}{A} and \tab{tab:sup:Mechanical-Elastomer}.
We choose to use EcoFlex~00-30 due to its Young's Modulus, density, and curing time, as well as the fact that it can withstand the degassing procedure.

\clearpage

\begin{table}[p]
\centering
\caption{Mechanical properties of different sensing surface material candidates.}
\resizebox{\textwidth}{!} {
\begin{tabular}[t]{cccccccccc}
\toprule
Material &$\rho $ [g/cm$^3$]&$V$ [cm$^3$] &$m$ [g] &$E$ [kPa] &$\nu $ & $A_e $ [\%] & Curing Time &Degas \\
\midrule
\textbf{EcoFlex 0030}  &\textbf{1.07}&\textbf{18.6}  &\textbf{23}  &\textbf{70}  &\textbf{0.49}  &\textbf{900} &\textbf{5\,h}&\textbf{Yes} \\
 EcoFlex 0035  &1.07 &18.6  &23  &70  &0.49  &900 & 10\,min &No \\
 EcoFlex 0050  &1.07 &18.6 &23  &80  &0.49  &980 & 4\,h &Yes\\
\bottomrule
\end{tabular}}
\label{tab:sup:Mechanical-Elastomer}
\end{table}%

\begin{table}[p]
\centering
\caption{Mechanical properties of different sensor skeleton material candidates.}
\resizebox{\textwidth}{!} {
\begin{tabular}[t]{ccccccccccc}
\toprule
Material &$\rho $ [g/cm$^3$]  &$V$ [cm$^3$] &$m$ [g] &$E$ [GPa] &$\nu $ & $A_e $ [\%] & $\sigma _y$ [Mpa] &$F$ [N] & $\sigma$ [Mpa] & $d$ [mm]\\
\midrule
 Steel  &7.86 &3.0 &23.6 &147  &0.30  &2.3  &455   &90 &433.1 &0.48\\
 \textbf{Aluminum}     &\textbf{2.68} &\textbf{3.0} &\textbf{8.0}  &\textbf{73}   &\textbf{0.33}  &\textbf{4.1}  &\textbf{227}   &\textbf{40} &\textbf{192.4} &\textbf{0.46}\\
 Tough  &1.20 &3.0 &3.6  &1.6  &(0.49)   &24.0   &60.6  &9  &50.44 &5.46\\
\bottomrule
\end{tabular}}
\label{tab:sup:Mechanical-Skeleton}
\end{table}%

\paragraph{Structure and material of the skeleton}
Considering weight, robustness, and yield strength, we design a beam structure as shown in \figS{fig:sup:mechanical}{B} and compare different materials for the skeleton in \tab{tab:sup:Mechanical-Skeleton}.
Several design choices are considered for the skeleton structure design.
From the functionality perspective, we mimic human fingertips when arranging the soft sensing areas over the surface, spaced out by the beams of the rigid frame.
From a mechanical perspective, the skeleton should be firm and durable enough to allow for high-force impact (collision).
At an early phase of the design of \method{}, we implemented finite element simulations to identify and reduce stress concentrations in the structure. We meshed the skeleton into small elements and applied a constant contact force with a magnitude of $40$\,N to each mesh point (normal to the surface at each point).
Our criterion was that the maximum von Mises-Stress should not exceed the elastic yield stress limit of the chosen material.
This analysis led to the particular chosen skeleton design with rounded edges and smooth connections between the beams; the beams have a circular cross-section with a diameter of 1.6\,mm.
The relevant results are summarized in \tab{tab:sup:Mechanical-Skeleton} with regard to material choices.
In the end, we chose the aluminum alloy due to its low weight.
Other designs are conceivable, either to increase the maximal force (above $40\,$N) or to change the sizes and shapes of the soft patches, which we leave to future work.

\paragraph{Over-molding}
A robust connection is required between the elastomer and the skeleton over which it is molded.
We test the minimal over-molding thickness of the elastomer as shown in \fig{fig:sup:mechanical}{C}.
We check whether the elastomer can cover a test cylinder without any defects.
We find a minimal thickness for a robust connection to be around $0.8\,$mm.

\paragraph{Relative positioning}
The relative position of the skeleton inside the elastomer additionally affects the system's sensitivity to contact.
We build a finite element model based on the material properties in \tab{tab:sup:Mechanical-Elastomer} and \tab{tab:sup:Mechanical-Skeleton} to analyze various possible relative positions between the skeleton and the elastomer.
The analysis shows that positioning the skeleton with an offset near the internal surface will increase the sensitivity because this relative positioning causes more displacement -- see \fig{fig:sup:mechanical}{D}.
In our design, the soft shell is $4\,$mm thick, and the skeleton is located $0.8$\,mm from the internal elastomer surface and $1.6\,$mm from the outer surface.

\subsection{Data interpretation}\label{sec:app:data-interpret}

\begin{figure}
    \includegraphics[width=\textwidth]{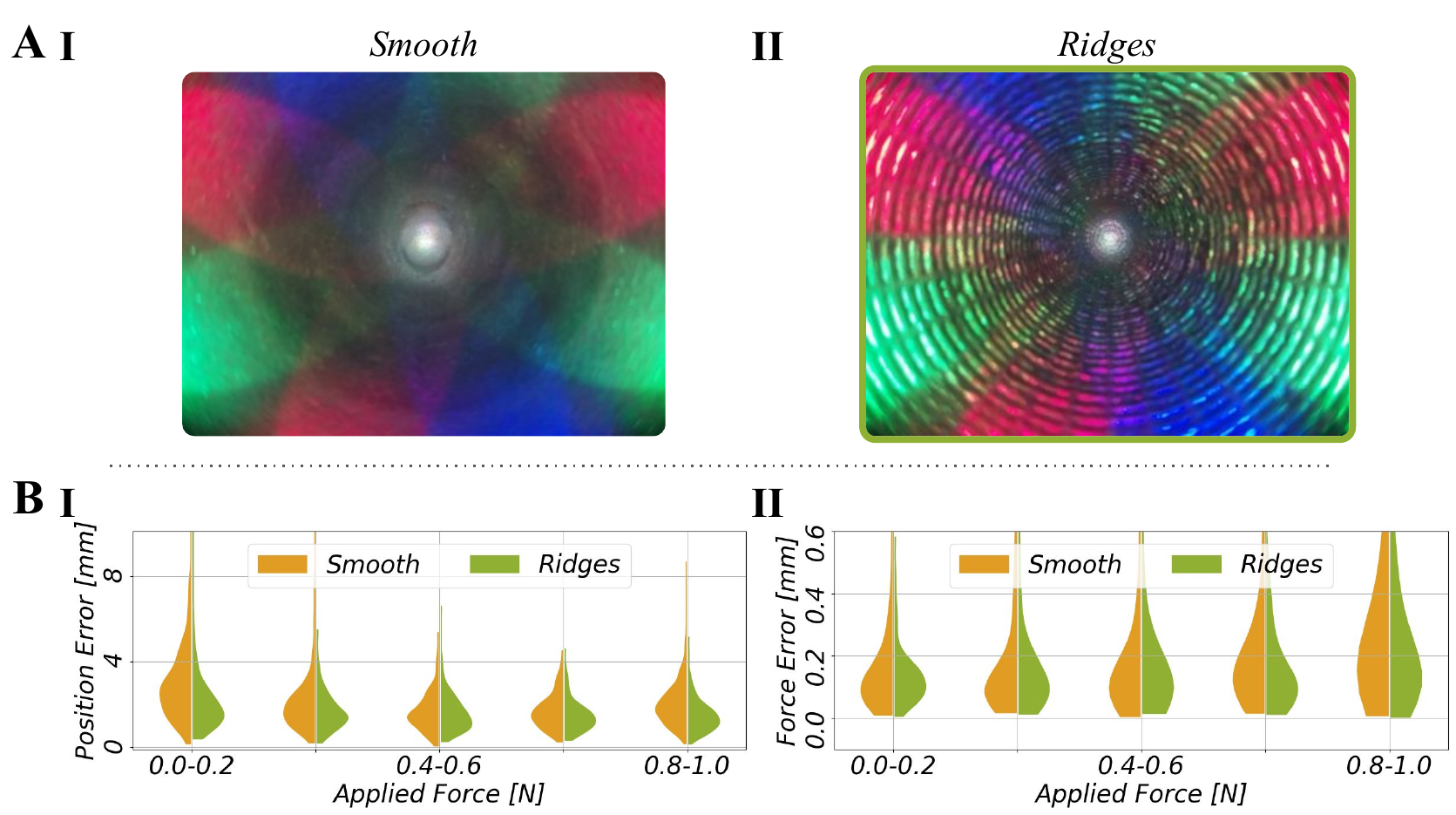}
    \caption{\textbf{Effect of the surface structure on the sensor performance}.
    \textbf{A} indicates two versions of internal surface morphology: one is a smooth surface, and another is with ridges.
    \textbf{B} compares the sensor performance of localizing single contact (\textbf{B-I}) and force quantification (\textbf{B-II}) with different force strengths, respectively. The errors in the case of the ridged surface (right half-violins) are generally lower than
     for the smooth surface (left half-violins).
    }
    \label{fig:sup:morphology}
\end{figure}
\paragraph{Surface morphology}
We investigate the effect of the surface morphology on the sensor performance.
As shown in \fig{fig:sup:morphology}, we compare a smooth inner surface to a surface with ridges.
Taking the single contact evaluation performance as the criterion, we find the surface with ridges  improves the localization and force quantification, as seen in \fig{fig:sup:morphology}{B}.
Moreover, we empirically find that the surface with ridges helps to accelerate the machine learning training procedure.
Furthermore, it is easier to track the movement of the ridges than that of the smooth surface; one should be able to design more advanced computer vision algorithms to improve sensor performance.

\paragraph{Indenter approximation}\label{sec:app:approximation}
To generate the force map data set, we need a force approximation model to describe the force distribution when the indenter is contacting \method{}'s outer surface.
We measure the total applied force vector with the force-torque sensor at the indenter (\figS{fig:Figure1}{D} part 2).
One theoretical approach to modeling how this force is distributed is to use Hertz contact theory~\cite{HertzTheory}.
For a spherical indenter with radius $\sigma=2$\,mm contacting an elastic half-space through a contact area with the same radius, this distribution is given by the following formula:
\begin{equation}
    F_H(x) =
    F \frac{1}{Z}\begin{cases}
        \left(1-\cfrac{x^2}{\sigma^2}\right)^{\frac{1}{2}}, & \text{if } |x|\leq \sigma\\
        0, & \text{otherwise}
    \end{cases}
    \label{eqn:sup:hertztheory}
\end{equation}
where $x$ is the radial distance from the contact center point, $F$ is the measured force, and $Z$ is the normalization constant such that the integral of the profile is $1$.

\begin{figure}
    \includegraphics[width=\textwidth]{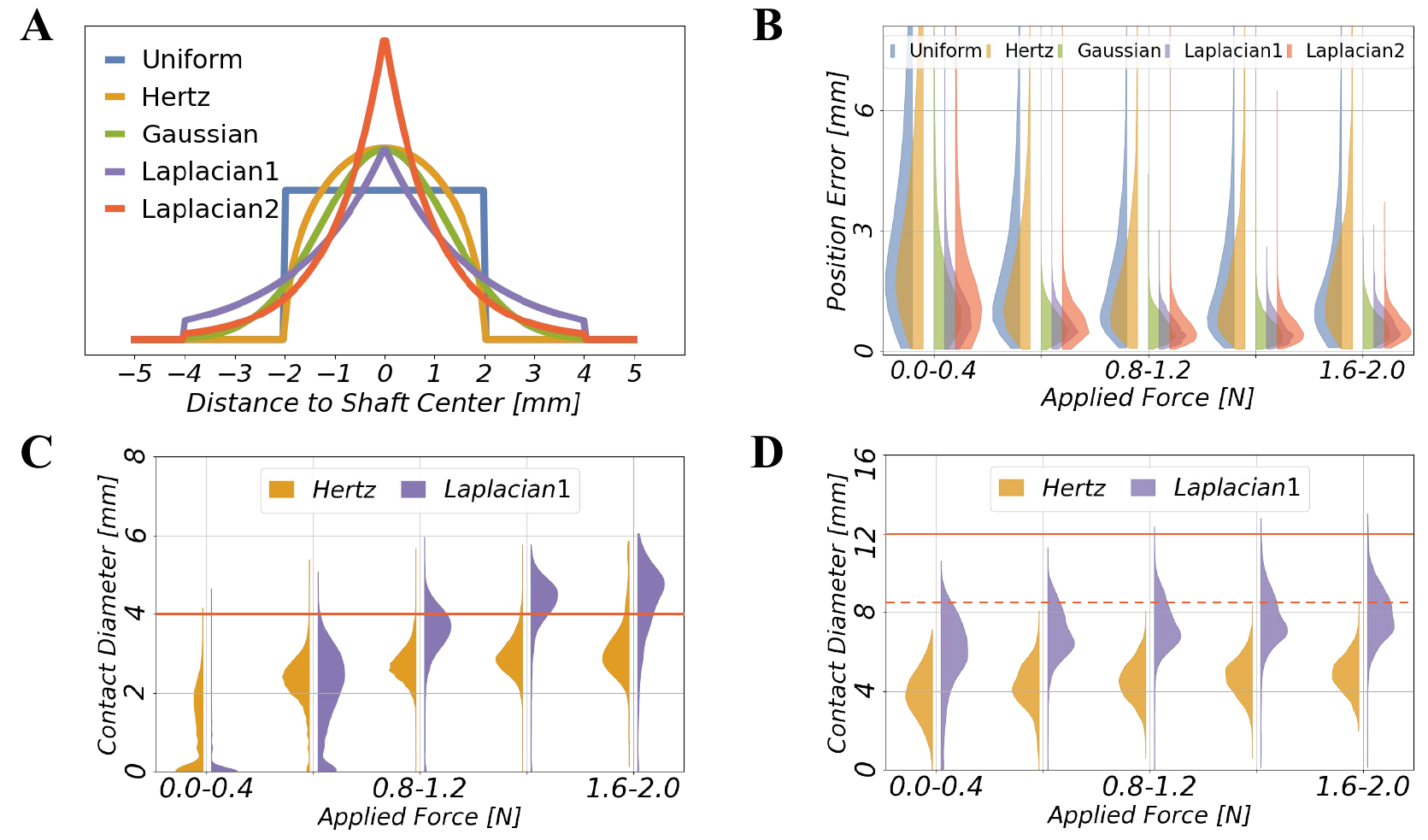}
    \caption{\textbf{Force distribution approximation and evaluation}.
    \textbf{A}: The five approximate force distribution methods that we tested for the $4\,$mm sphere-shaped indenter.
    The illustration is in one dimension and was revolved to distribute the measured contact force vector across the local surface of the force map around the contact point.
    The uniform (blue) and Hertz theory (orange) distribution curves are strongly localized with a radius of $2\,$mm. 
    The green curve follows a truncated Gaussian distribution, and the two other curves (purple and red) follow Laplacian distributions that also stop at a radius of $4$\,mm.
    \textbf{B} shows the position inference performance using the different methods.
    \textbf{C} and \textbf{D}: The diameters of the contact area prediction for two indenter sizes ($4\,$mm, and $12\,$mm) using two different approximation methods (Hertz and Laplacian1).
    The dashed line indicates a reference diameter from \fig{fig:sup:largerindenter}{A-I} based on the fact that the large indenter penetrates into the sensor only partially.}
    \label{fig:sup:approximation}
\end{figure}

\begin{table}
\centering
\caption{Evaluation of force distribution approximation. We examined the performance of five different force distribution approximation curves for the $4\,$mm indenter. The first three presented values are median errors of different types. The two values in the ``Contact Diameter'' column show the median contact area size based on two force thresholds for counting contact points; the first diameter uses a force threshold of $0.02\,$N, and the second uses $0.01\,$N.}
\resizebox{\textwidth}{!} {
\begin{tabular}[t]{ccccccccccc}
\toprule
Shape     & Position Error [mm] & Force Error [N] & Angle Error [$^\circ$] & Contact Diameter [mm]\\
\midrule
Uniform    &1.9   &0.10   &11.1   &2.4, 2.9\\
Hertz      &1.8   &0.11   &12.4   &2.4, 2.7\\
Gaussian   &0.7   &0.10   &11.7   &2.6, 3.7\\
\textbf{Laplacian1} &\textbf{0.6}   &\textbf{0.08}   &\textbf{10.2}   &\textbf{2.4, 4.2}\\
Laplacian2 &0.7   &0.12   &13.7   &2.2, 3.5\\
\bottomrule
\end{tabular}}
\label{tab:sup:approximation}
\end{table}%
However, Hertz theory is not appropriate for thin elastic sheets that deform as a whole under the force of the indenter.
More importantly, the Hertz profile causes problems in our machine-learning procedure because it strongly localizes the target signal.
We have a mesh grid of the sensor's outer surface, and this grid has $3\,800$ points with neighboring points separated by around $1\,$mm.
The indenter has a diameter of $4\,$mm.
Using the Hertz profile causes at most $13$ points among $3\,800$ points to have a non-zero value, which we found empirically to be hard to train using the machine-learning procedure.
Thus, we check four alternative profiles to understand this issue and verify the benefits of distributing the force more widely, as illustrated in \fig{fig:sup:approximation}.
First we consider a Laplacian profile with a cutoff at 2$\sigma$:
\begin{equation}
    F_{L}(x) = F \frac{1}{Z}
    \begin{cases}
        e^{- \frac{|x|}{\lambda}}, & \text{if } |x|\leq 2\sigma\\
        0, & \text{otherwise}
    \end{cases}
    \label{eqn:sup:approximation}
\end{equation}
and two shapes: Laplacian 1 with $\lambda=0.87\sigma$, which has the same maximal value as the Hertz model,
and Laplacian 2 with $\lambda=0.5 \sigma$, which is more peaked.
Another alternative is a truncated Gaussian distribution:
\begin{equation}
    F_G(x) =
    F \frac{1}{Z}\begin{cases}
        e^{\left(-\frac{1}{2}\cdot\frac{x^2}{0.4\sigma^2}\right)}, & \text{if } |x|\leq 2\sigma\\
        0, & \text{otherwise.}
    \end{cases}
    \label{eqn:sup:gaussian}
\end{equation}
Finally, we tested a uniform distribution with the indenter's radius as a reference:
\begin{equation}
    F_U(x) =
    F \frac{1}{Z}\begin{cases}
        1, & \text{if } |x|\leq \sigma\\
        0, & \text{otherwise.}
    \end{cases}
    \label{eqn:sup:uniform}
\end{equation}

A comparison of the different models and the resulting performance is shown in \fig{fig:sup:approximation} and \tab{tab:sup:approximation}.
We find that the approximation maps with a smoother profile (Laplacian and Gaussian) achieve better position accuracy than those with sharp edges.
In particular, the Laplacian 1 with its flatter profile yields the best accuracy in position, force magnitude, and force direction prediction.
All approximations tend to have similar contact area prediction, with a slight overestimation for the smoother profiles for strong forces.
Based on these results, we choose \textit{Laplacian 1} (\eqn{eqn:sup:approximation} with $\lambda=0.87\sigma$) as our force-distribution approximation method.

Our interpretation of why the smoother profiles achieve better results is that they cover a larger neighborhood and therefore reduce the sparsity in the target signals.
The machine-learning model has the tendency to produce a smoothed output, so the peak of our approximation model is smoothed out.

An alternative to the approximation would have been a finite-element method to compute the local force distribution.
We did not pursue such an approach for two reasons. First, with our simulation tools the simulation of all tested contact locations would take around 50 days.
Second, the linear assumption between deformation and force in the simulation is violated in our sensor design due to large deformations.

\begin{figure}
    \vspace{-2cm}
    \includegraphics[width=\textwidth]{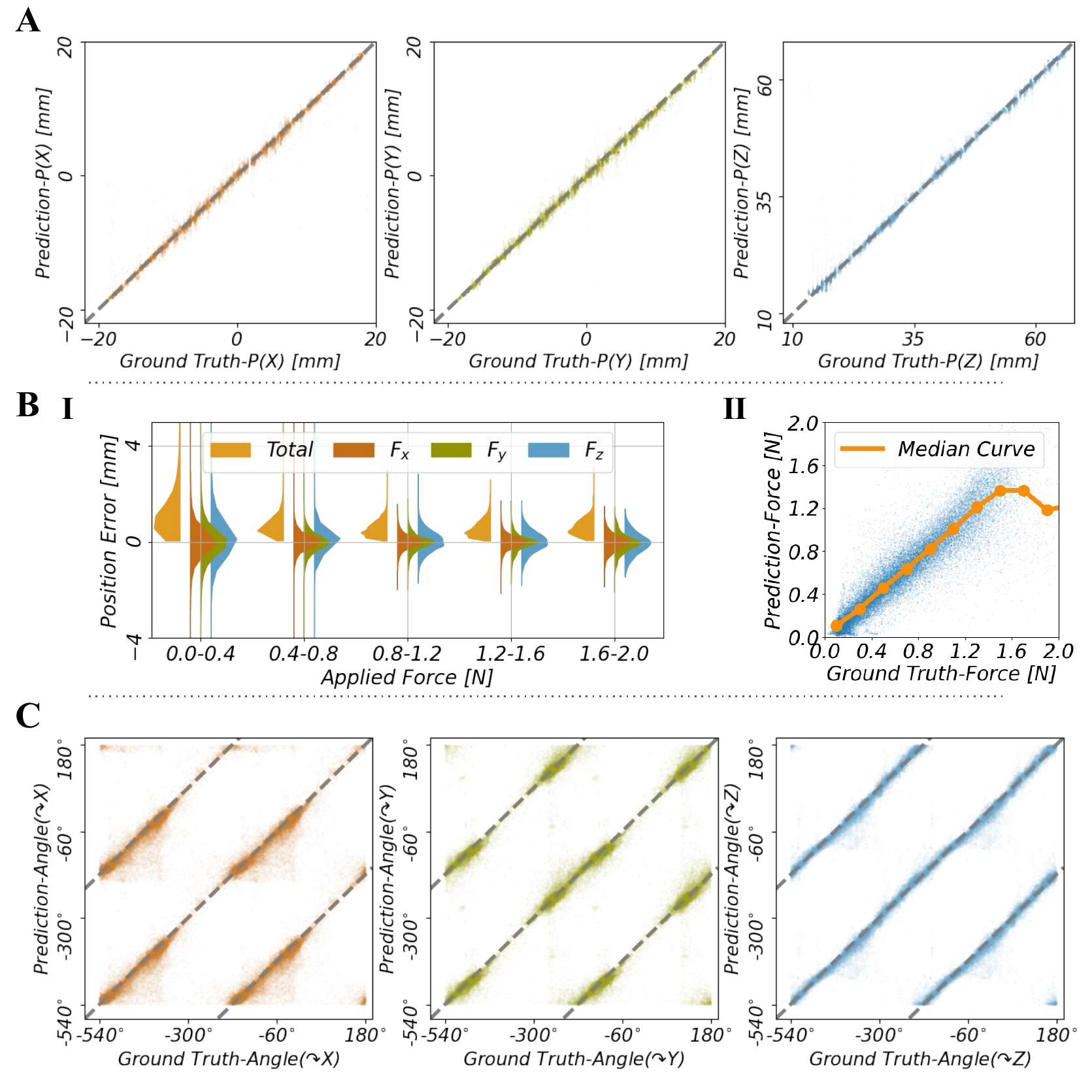}
    \caption{\textbf{Force map evaluation}.
    Based on the estimated force map, we extract information about contact position (\textbf{A}, \textbf{B-I}), force amplitude (\textbf{B-II}), and force direction (\textbf{C}).
    \textbf{A} shows the ground truth and estimated positions in $x, y, z$ directions, where
    \textbf{B} contains the corresponding statistical evaluation.
    \textbf{B-I} presents the localization performance grouped by applied force strength.
    The red, green, and blue colored half-violins show the distribution of deviations in the $x$, $y$ and $z$ directions, respectively.
    The orange half-violins are the resulting total errors.
    \textbf{B-II} shows the estimated force magnitude as a function of the ground truth force. The median curve indicates a good overall correspondence with a tendency toward underestimation for larger forces, partially caused by a paucity of data in this regime.
    \textbf{C} presents the estimated force directions indicated by the angles around the $x$, $y$, and $z$ axes.
    The gray dotted lines indicate perfect prediction.
    For better visualization, and to avoid cropping, the angle range is extended to ($-540^\circ$ to $180^\circ$).}
    \label{fig:sup:accuracy}
\end{figure}

\paragraph{Force map evaluation}
Using the force map as the target output for our training data allows us to predict the force distribution directly from an image.
The performance of the system is summarized in \fig{fig:Figure4} in terms of force quantification and force direction estimation.
\fig{fig:sup:accuracy} presents a more detailed evaluation of the performance.
First, we want to quantify single-contact precision.
We select the $20$ points from the force map with the highest predicted force magnitudes and take the mean of their positions.
The localization performance shows no visible differences in the $x$, $y$, or $z$ directions; see \figS{fig:sup:accuracy}{A \& B-I}.
However, we get slightly worse results than our direct estimation approach (\fig{fig:Figure3}. Direct prediction has a median error of $0.4$\,mm and $0.03\,$N, whereas the inference computed from the force field has a median error of $0.6$\,mm and $0.08$\,N.

Second, we consider the force magnitude, which is displayed in \figS{fig:sup:accuracy}{B-II}.
The estimated force tends to under-estimate the actual forces for strong applied forces.

Third, we present \method{}'s performance in estimating the force direction.
\FigS{fig:sup:accuracy}{C} shows the true and predicted force angles relative to the three coordinate axes $x$, $y$, and $z$ (the axes are shown, \eg, in \fig{fig:sup:multiple}).
Overall, we observe a very good correspondence.
The sensor can estimate the rotation direction around the $z$-axis better than the other two, while the $x$ and $y$ directions show similar performance.

\paragraph{Force map evaluation for a larger indenter}
\label{sec:sup:largerindenter}
We quantitatively validate the ability of our machine-learning pipeline to generalize the force map inference problem by conducting experiments with a larger indenter. The same test bed was used to probe \method along its entire outer surface with a hemispherical indenter with a diameter of $12\,$mm, which is three times wider than the indenter with which the training data set was collected. We then process the data through the previously trained machine learning model and compare it with the ground truth.
\begin{figure}
    \vspace{-2cm}
    \centering
    \includegraphics[width=0.9\textwidth]{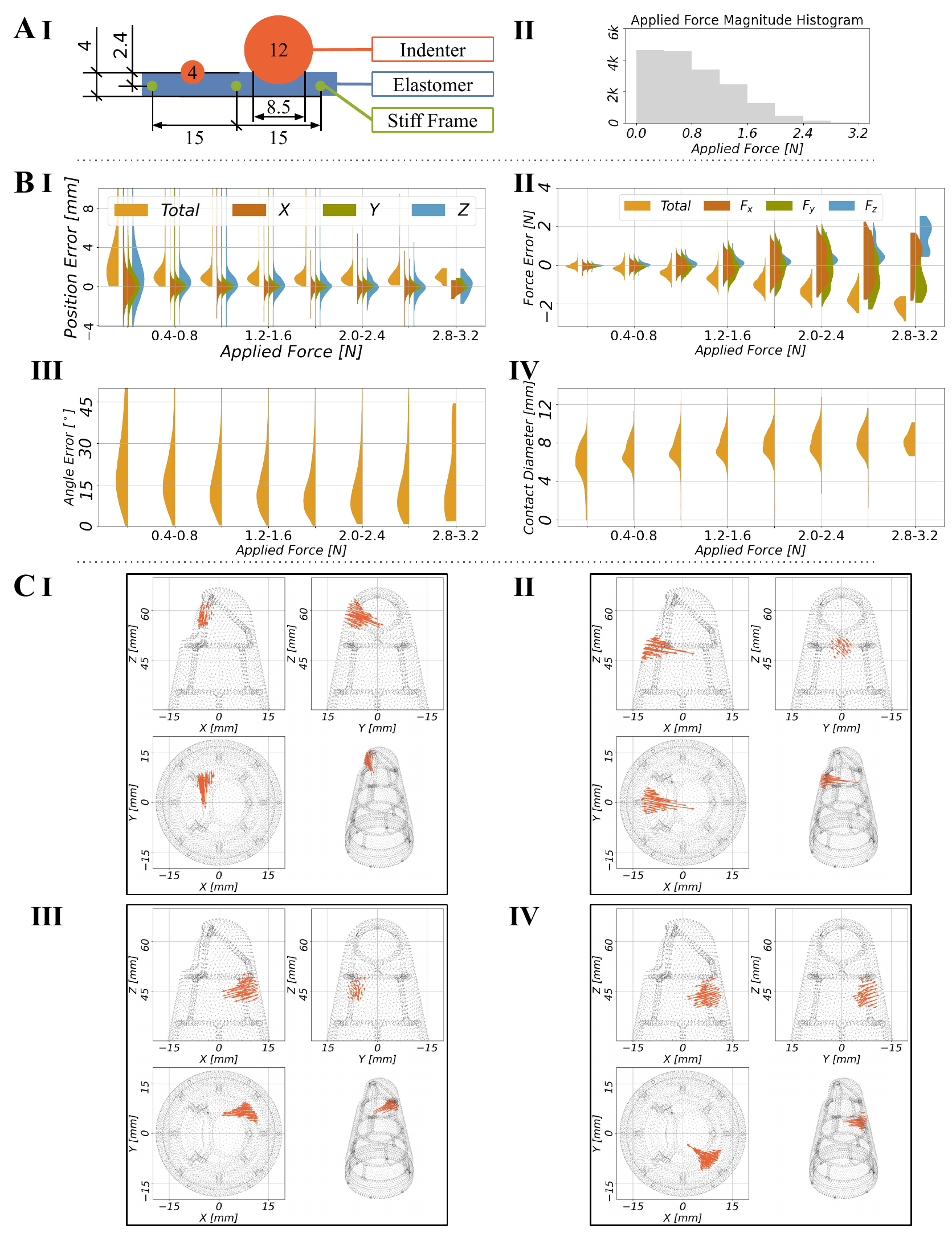}
    \caption{\textbf{Force map evaluation with a larger indenter}.
    An indenter with $12\,$mm diameter is used to validate the sensor performance.
    \textbf{A-I} shows the sizes and positions of the relevant components; \textbf{A-II} shows a histogram of the applied force.
    \textbf{B} presents an evaluation of test accuracy for localization, force strength, force direction, and contact area diameter.
    \textbf{C} demonstrates four sample cases of the force map prediction. C-I--III show an asymmetric force prediction that is plausible because the indenter is hitting a beam. C-IV shows the case without the beam, and the prediction is symmetric, as expected.}
    \label{fig:sup:largerindenter}
\end{figure}

The geometry of the indenters, the elastomer, and the stiff frame are visualized in \fig{fig:sup:largerindenter}{A-I}.
The evaluation is on $16\,919$ indentation samples with widely varying applied force magnitudes, as shown in the histogram of \fig{fig:sup:largerindenter}{A-II}.
The force threshold for the data collection is set to $3\,$N, but almost all collected data points have a total force less than $2\,$N.
This pattern indicates that at some positions the large indenter experiences a sudden force increase, quickly raising the total resultant force over the threshold.
This situation happens when the indenter hits the stiff frame directly.
Given the lack of data at higher forces, we evaluate the sensor performance with an applied force threshold of $2\,$N.

\fig{fig:sup:largerindenter}{B} shows the results of this quantitative evaluation.
The median position error is $1\,$mm; it is largest ($1.6\,$mm) at smaller indentation forces, and it gets smaller ($0.8\,$mm) with higher indentation forces.
The median force error is $0.23\,$N; it is smallest ($0.09\,$N) at smaller indentation forces, and it gets bigger ($1\,$N) with higher indentation forces.
The median direction error of the force is $16^\circ$; it is largest ($16^\circ$) at smaller indentation forces and gets smaller ($9^\circ$) with higher indentation forces.
The median estimated contact area diameter increases from $6\,$mm to $8.5\,$mm as the contact force increases, which approximates Hertzian contact theory for a large indenter with relatively small indentations.

In addition, we include a qualitative evaluation on the force map prediction as shown in \fig{fig:sup:largerindenter}{C}.
When the indenter presses on the soft material far away from the stiff frame, the force map shows a symmetric force distribution.
When the indenter contacts an area near the stiff frame, it shows an asymmetric force distribution.
To be more explicit, the strongest estimated contact force vectors appear directly at the frame, and smaller force vectors occur nearby, as would be expected from an inhomogeneous surface like this.

\paragraph{Sensitivity: sensor posture recognition}
A haptic sensor created from soft materials also deforms due to gravity and inertial effects.
In some contexts, these deformations will be considered disturbances that should be ignored in favor of contact signals.
In other contexts, we can make use of these effects to enhance the capabilities of the sensor.
Here, we demonstrate that gravity causes deformations that can be captured by \method{}'s camera, and the sensor's posture can be recognized.
To quantify the effect, we can train a machine-learning model to estimate the orientation of the sensor in terms of roll and yaw angles, as depicted in \figS{fig:Figure5}{A-I \& II}.

In a quasi-static scenario, the sensor is held in the air with varying yaw angles (from $-90^\circ$ to $90^\circ$) and roll angles (from $0^\circ$ to $360^\circ$). Even though our elastomer skin is very opaque, we recorded the data for this experiment in a dark laboratory at night, to avoid any potential artifacts from external illumination.
The raw difference between the captured images has a root-mean-square (RMS) difference per pixel of $0.012$ due to noise and maximally $0.016$ due to gravity, where $1$ is the maximum possible (see \figS{fig:Figure5}{A-III}).
The estimation performance of the machine-learning model is provided in \figS{fig:Figure5}{A-IV}.
Even from the slight deformation caused by gravity, the self-posture can be estimated with an overall mean accuracy of $2.5^\circ$ in the yaw direction and $11.6^\circ$ in the roll direction.
The high error of estimating the roll angle appears when the roll axis aligns with the gravity vector, as expected.

Despite the sensor's ability to estimate its own posture, the gravitational effect is still very small in practice and does not significantly affect the sensor's main goal of perceiving external physical contacts, which are generally much larger than the self-weight of the elastomer skin.
\vid{S6} shows the system accurately detecting contacts in different postures.

\paragraph{Tactile fovea}
Our current version of \method{} possesses a nail-shaped zone with a thinner elastomer layer, as indicated in \figS{fig:Figure5}{B}; with a sensing area of $13\times11\,$mm$^2$, this tactile fovea is designed for detecting tiny forces and perceiving detailed object shapes.
Based on FEA results and real experiments, we find that the thicker elastomer layer on the rest of the stiff skeleton smooths the shape of the contacted object, so that it is not easy to detect the exact shape of small objects.
A very thin elastomer layer would be ideal but is also too fragile for vigorous interaction.
We balance these two effects and choose a thickness of $1.2\,$mm for this special sensing zone.
In the Results section we report its higher position accuracy and force accuracy in comparison to  the other sensing areas.

We conduct two demonstrations of how the fovea could be used for shape detection.
The first one represents a v-shaped wedge with different levels of sharpness (included angles from $10^\circ$ to $180^\circ$); see \figS{fig:Figure5}{B-III} and \fig{fig:sup:sensitivity}.
The second set of samples represents extruded polygons with an increasing number of edges (triangle, square, pentagon, hexagon, etc.); see \figS{fig:Figure5}{B-IV} and \fig{fig:sup:sensitivity}.
\Fig{fig:Figure5}{B-II} shows that \method{}'s camera can visually distinguish the tested samples up to $150^\circ$ and nine edges. An automatic procedure is left for future work.
Theoretically, if the indentation depth increases, the shape detection accuracy will be improved.
However, to not destroy the sensor by exceeding the elastomer's elongation capabilities, we limit the max indentation depth to $18\,$mm, which corresponds to half of the maximal deformation; the maximum resulting net force is $1.2$\,N in this area.

\subsection{Machine Learning Details}\label{sec:app:ml-details}
The raw images are interpreted into understandable haptic information using a custom machine-learning method.
To tell when, where, and how the sensor is contacted, we present two formats to quantify sensor performances: direct single-contact inference with location and resultant directional (normal/shear) force, and three-dimensional force map over the 3D conical surface.
In addition, the sensor posture is also inferred through a machine-learning method.

Our three information formats are trained using the same machine-learning-based architecture \textbf{ResNet-18}~\cite{ResNet} but with customized modifications.
The original ResNet-18 has one input convolution layer, one pooling layer to adjust the input size, and four standardized ResNet blocks afterwards connecting with a fully connected output layer. In all models we use an input image size of $410\times308$, which is down-sampled from the $1640\times1232$ image captured by the camera; the number of channels depends on the task. Further details are provided in the following paragraphs specific to each inference task.

\paragraph{Direct single-contact inference}
We use the standard ResNet-18 architecture with six input channels: three for the RGB difference between the input image and the reference image, and three for a static RGB image of the skeleton before over-molding (see \fig{fig:Figure3}).
The fully connected output layer predicts six channels: the three positions ($x,y,z$) where it estimates the contact is occurring, and the three force components (shear forces and normal force relative to the surface) at that location.

\paragraph{Force map inference}
For the force map inference, we modify the architecture.
The input is the same as for the direct contact inference.
However, we only use two ResNet blocks instead of four and replace the fully connected output by the following set of operations:
\begin{enumerate}[itemsep=0em]
    \item An appropriate up-scaling convolution using a transposed convolution to obtain the output dimension $64\times64$ from the $52\times 39$ output of the second ResNet block \\
    \resizebox{.93\textwidth}{!}{\tt ConvTranspose2d(128, 128, kernel\_size=(7, 7), stride=(1, 1), dilation=(4, 2), output\_padding=(1, 0)))}
    \item A convolution layer from 128 to 64 channels\\
    {\tt\scriptsize Conv2d(128, 64, kernel\_size=(3, 3), stride=(1, 1), padding=(1, 1))}
    \item Batch normalization followed by a LeakyReLU
    \item A convolution layer to the force map ($64\times 64$ with three channels) \\
    {\tt \scriptsize Conv2d(64, 3, kernel\_size=(3, 3), stride=(1, 1), padding=(1, 1))}
\end{enumerate}
There are several reasons for this architecture: the output is a spatial map where each output pixel is mostly influenced by a local region in the input.
Thus, maintaining a good spatial resolution is advantageous. After two ResNet blocks, we still have a resolution of $52\times 39$, which is similar to our desired output.
Every additional block reduces the dimensions and adds considerable computation.
The network has in total 1.5 million parameters.
We are convinced that a smaller architecture can be found, but it was not the aim of our research.

\paragraph{Sensor posture inference}
For the task of predicting the posture of the sensor, we also use the standard ResNet-18 architecture. The input is the three channels of the difference image, and the outputs are the yaw and roll angle.

\subsection{Functionality illustration}\label{sec:app:function}
\begin{figure}
    \includegraphics[width=\textwidth]{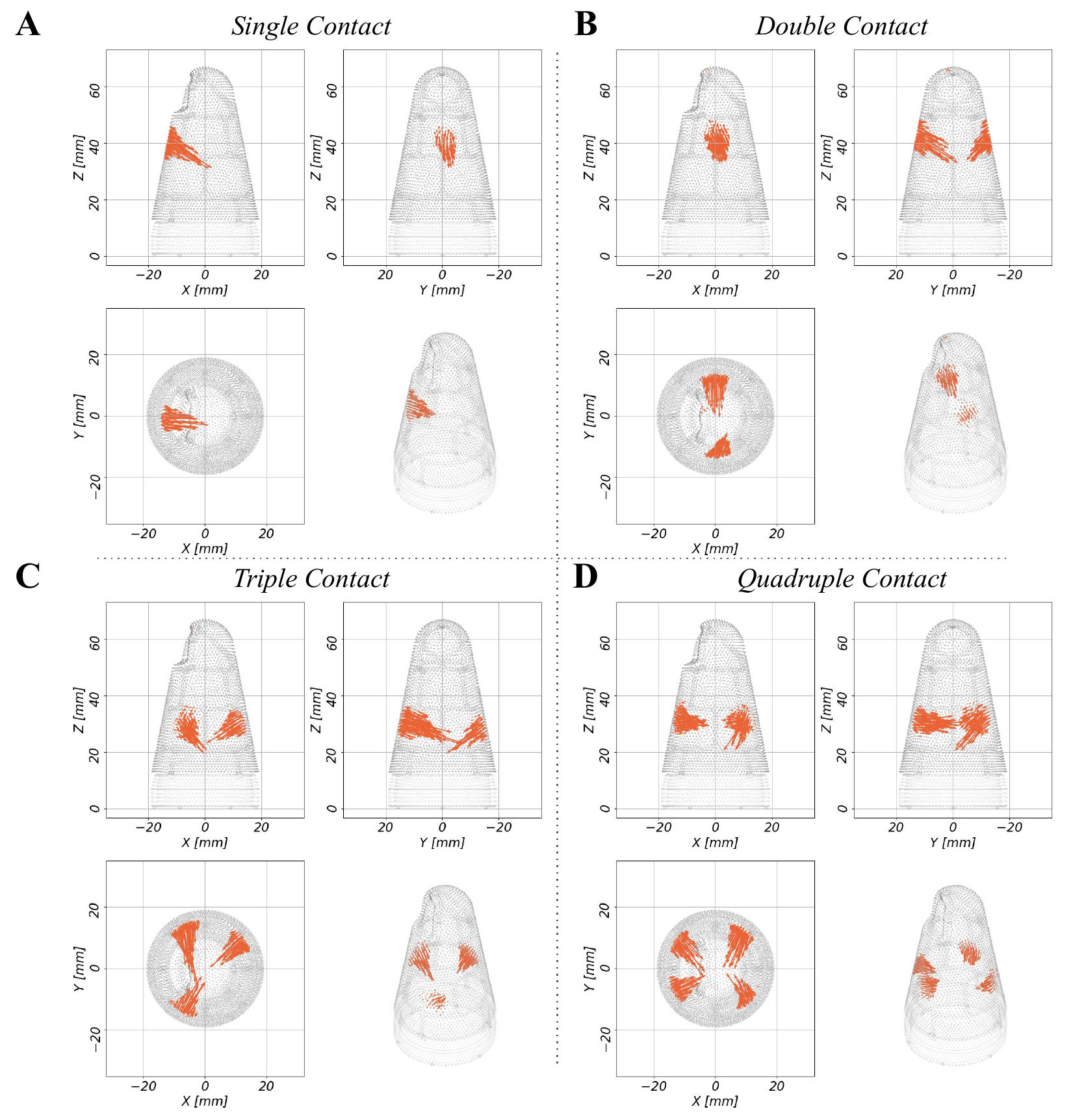}
    \caption{\textbf{Multiple contacts examples}.
    Visualizations of the force map distributions over the sensing surface for a single contact (\textbf{A}), double contact (\textbf{B}), triple contact (\textbf{C}), and quadruple contact~(\textbf{D}).}
    \label{fig:sup:multiple}
\end{figure}

\begin{figure}
    \includegraphics[width=\textwidth]{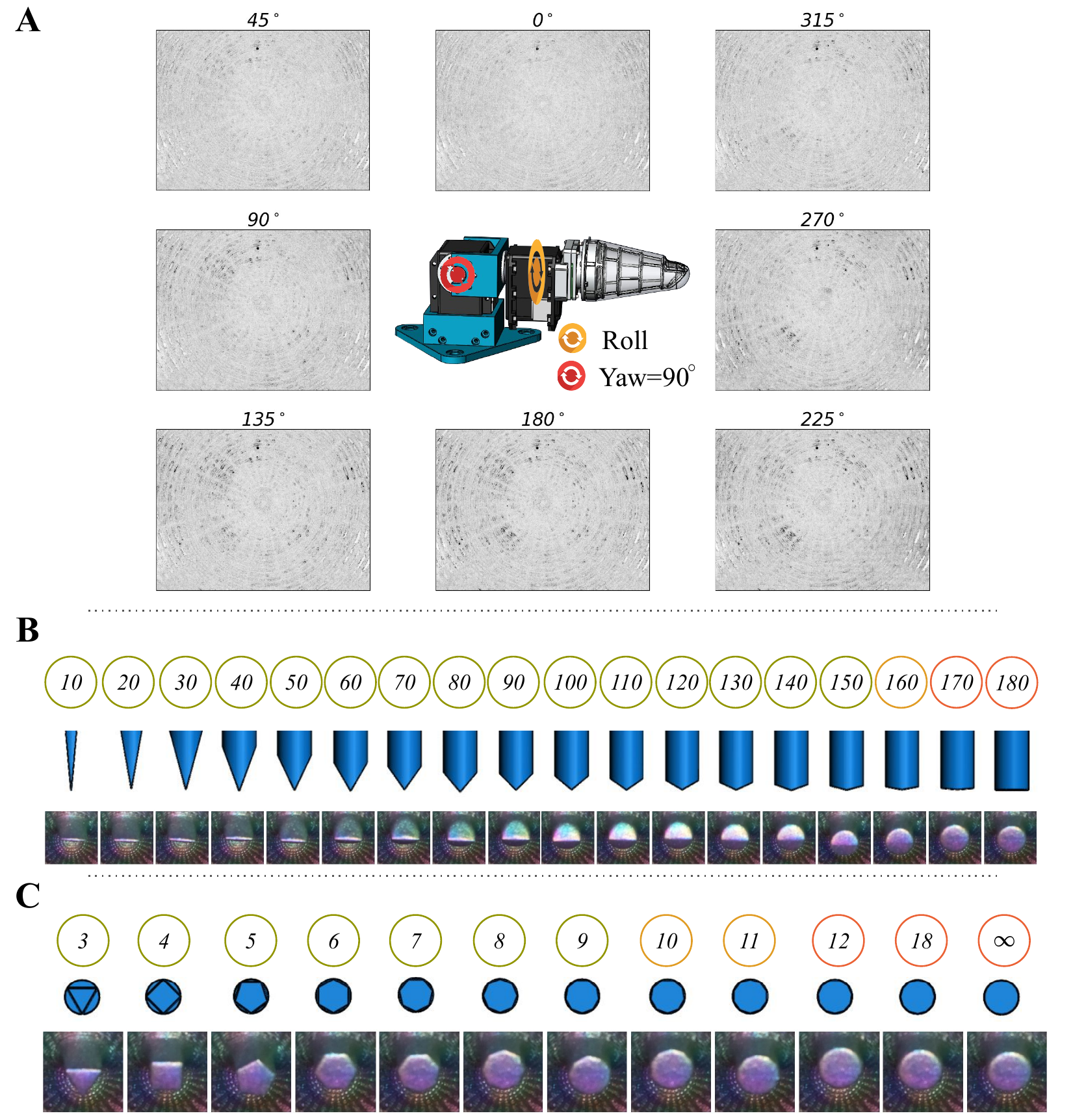}
    \caption{\textbf{Sensitivity evaluation}.
    \textbf{A} shows the image changes caused by gravity when the sensor rotates $360^\circ$ around the roll direction while maintaining a yaw angle of $90^\circ$.
    In contrast to the images actually used for the posture detection experiment, the images presented here were recorded in typical overhead lighting conditions. Nevertheless we see no illumination impact even at the thin fovea part, showing the skin is sufficiently opaque.
    \textbf{B} and \textbf{C} extend the reported evaluation of the sensitivity of shape detection for wedge sharpness and polygon edges.}
    \label{fig:sup:sensitivity}
\end{figure}

\paragraph{Static evaluation}
We qualitatively show the performance of multiple contacts in \fig{fig:sup:multiple}, the sensitivity of recognizing self-posture in \fig{fig:sup:sensitivity}{A}, and shape detection in terms of the sharpness of a V-shaped wedge (\fig{fig:sup:sensitivity}{B}) and the number of edges in a polygon (\fig{fig:sup:sensitivity}{C}).


\method{} can theoretically discriminate 16 contacts simultaneously, which coincides with the number of hollow areas formed by the skeleton.
\fig{fig:sup:multiple} shows four sample interactions where the sensor detects up to four contacts near one another on its surface.

Due to the softness of the elastomer, gravity and inertial effects can cause interference with the sensor's output.
We test the gravity effect and find that we can use the deformations caused by gravity to estimate the posture of the sensor.
As shown in \fig{fig:sup:sensitivity}{A}, the sensor is at the yaw angle of $90^\circ$ and is rotating along the roll direction.
The eight images show the difference between the raw image at different roll angles, plus a reference image at the roll angle of $0^\circ$.
The image at $0^\circ$ is not empty due to the noise of the imaging system.

\Fig{fig:sup:sensitivity}{B} and \fig{fig:sup:sensitivity}{C} show all test samples used to evaluate the shape detection performance, which supplements \fig{fig:Figure5}{B}.
By looking at these images by eye, we can visually discriminate V-shaped wedge sharpness up to $150^\circ$ and about $9$ or $10$ polygon edges.

\paragraph{Dynamic evaluation}
We quantitatively show the sensor's performance at localizing an indenter in sliding motion.
The experimental setup appears in \vid{S8}:
the indenter first contacts \method{} and then slides along the sensor surface for 4\,mm before stopping for 5\,seconds.
This sliding-and-stopping behavior is repeated five times in the forward direction and five times in the backward direction.
We use two complete cycles of this behavior.
After this, the indenter slides along the sensor in the forward direction and the backward direction without any pauses for another two cycles.

\begin{figure}[p!]
    \includegraphics[width=\textwidth]{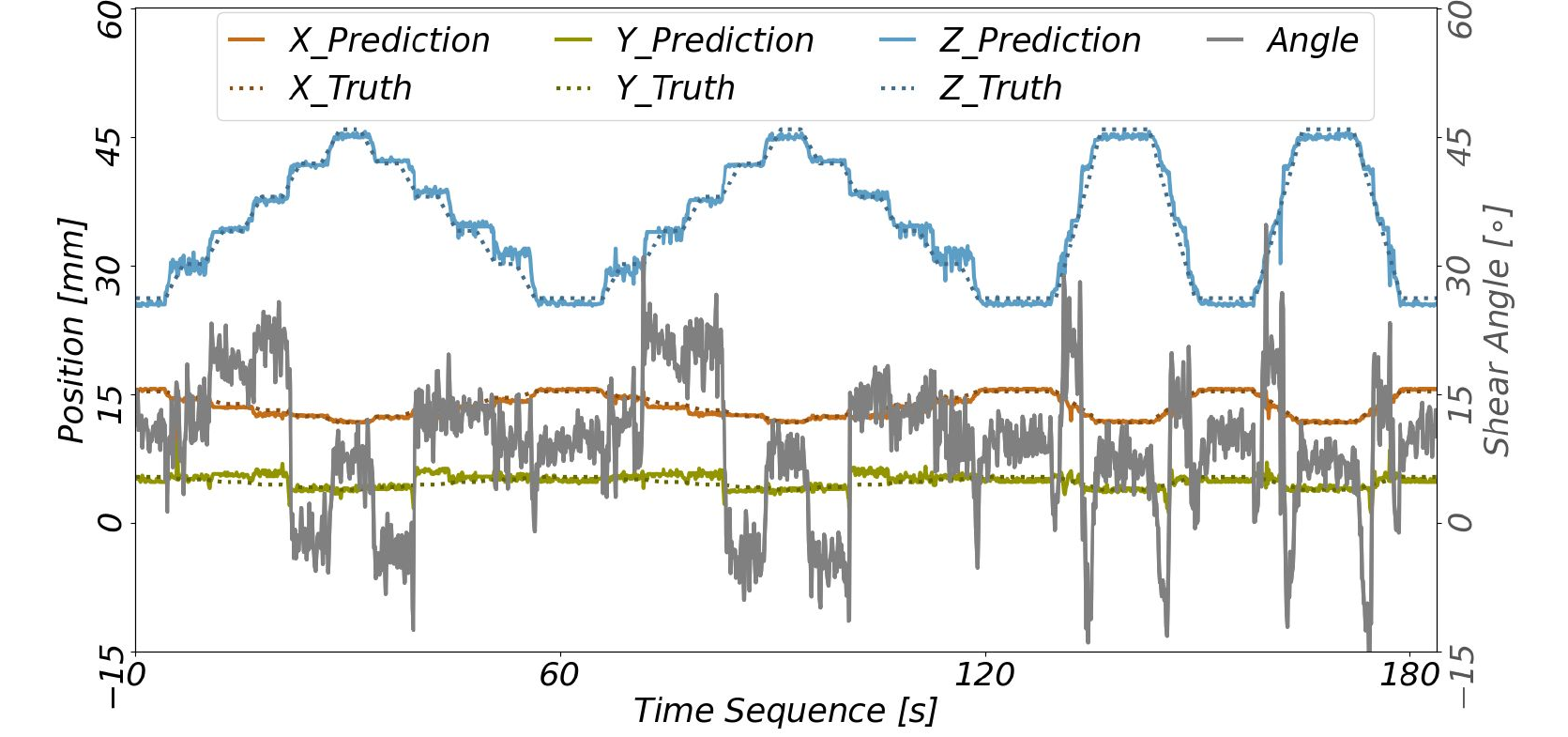}
    \caption{\textbf{Localizing the indenter in sliding motion}.
    This plot shows the sensor's capability of localizing an indenter when it is sliding along the sensor surface.
    The dashed lines show the actual position of the indenter over time, and the solid lines are the estimated contact locations.
    The listed axes refer to the global coordinate frame of \method{}.
    The gray line shows how the force angle changes during the sliding motion.
    It is the angle between the estimated force vector and the sensing surface normal vector.}
    \label{fig:sup:slide motion}
\end{figure}

The contact can be accurately localized, and the direction of the indented force can be discriminated as shown in \fig{fig:sup:slide motion}.
We evaluate the changes of the angle between the estimated force vector and the sensing surface normal vector during the sliding motion.
The gray line in \fig{fig:sup:slide motion} shows the angle change at each beginning of sliding motion with a recovering phase during the pause interval.
Between the sliding segments, there is one position that shows an abnormal angle change; it is caused by the metal beam of the skeleton.

\subsection{Ablation Studies}\label{sec:app:ablation}
\begin{figure}[p!]
    \centering
    \includegraphics[width=0.9\textwidth]{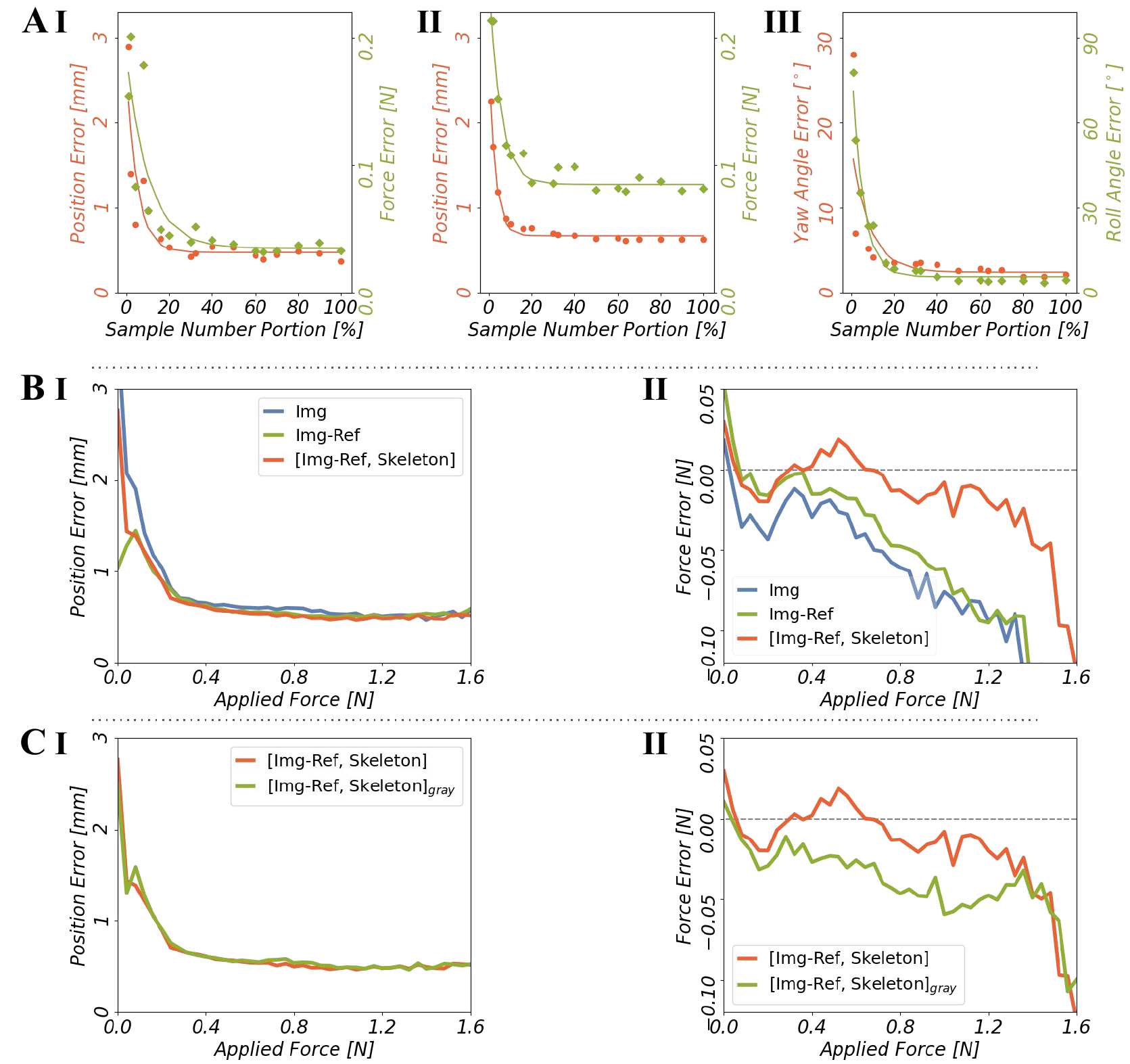}
    \caption{\textbf{Ablation studies on dataset size and network input}.
    \textbf{A} shows the ablation study on dataset size.
    \textbf{I, II} show the median position error (red) and force error (green) on the test set after training on different randomly sampled portions of the training dataset (about 112\,000 examples in total). The results are shown as dots (position error) and squares (force error) with fitted curves for illustration.
    \textbf{I} is the evaluation for direct single-contact inference, and \textbf{II} is for force-map inference.
    \textbf{III} shows the median of the sensor posture error on the test set after training on different randomly sampled portions of the training dataset.
    \textbf{B, C} show the ablation study on the network input.
    \textbf{B} shows the evaluation on the influence of different inputs on the force map prediction. We compare the original input [Img$-$Ref, Skeleton] (using the reference image subtracted from the raw image, as well as the skeleton) with only the raw image (Img) and the reference-subtracted image (Img$-$Ref).
    \textbf{C} presents the comparison between the original input ([Img$-$Ref, Skeleton]) and its grayscale version ([Img$-$Ref, Skeleton]$_{gray}$).
    }
    \label{fig:sup:ablation}
\end{figure}

\paragraph{Dataset size}
Machine-learning-driven sensors rely on copious data to train good models for real-world applications.
For our study, we design an automatic test bed to collect massive data for the three machine-learning models: direct force prediction, force-map prediction, and posture prediction.
Here, we analyze how many samples are needed to achieve good performance on these three learning tasks.
As shown in~\figS{fig:sup:ablation}{A-I}, for the direct single contact prediction, training with only 20\%--40\% of the data samples of the original training set ($\approx$ 112\,k) yields similar performance.
Similar observations are valid in the force map prediction and the self-posture inference, as shown in~\figS{fig:sup:ablation}{A-II, A-III}. We see two ways to further reduce the amount of required real-world samples: one is to find a machine-learning model that generalizes better from less data. The other is to augment the real data by simulated data and then use transfer learning, \eg~\cite{HapDefX}.

\paragraph{Network input}
Our method provides several inputs to the neural network to enable it to make accurate predictions. In order to understand the respective importance of these components, we provide ablations and modifications of this input.
In principle we have three sources of information: the raw captured image (Img), the reference image (Ref), and the skeleton image (Skel), which are combined to create the actual input as [Img - Ref, Skel]. We ablate this chosen design by providing only the raw image (Img) and by only subtracting the reference from the raw image (Img-Ref).
As shown in~\figS{fig:sup:ablation}{B}, using only the raw image shows the worst performance.
Subtracting the reference image slightly improves the performance in both localization and force quantification, but it suffers from force underestimation mostly at the locations of the skeleton.
Providing the network with the skeleton image seems to largely alleviate this underestimation problem.
One interpretation of these findings is that the skeleton image allows the network to readily distinguish the positions with the stiffer material and adjust the processing accordingly.

We also do an ablation study to analyze the effect of the colored structured light by using only grayscale images, which can be seen as only using photometric stereo (PS).
As shown in~\figS{fig:sup:ablation}{C}, removing the color from the structured light yields lower force accuracy.
In comparison, including the PS effect in our design achieves an average localization accuracy of 0.6 mm, one order of magnitude better than that achieved by GelTip~\cite{GelTip} (5\,mm).

\end{document}